\documentclass[preprint]{elsarticle}

\usepackage{hyperref}
\usepackage{url}
\usepackage[utf8]{inputenc} 
\usepackage[T1]{fontenc}    
\usepackage{hyperref}       
\usepackage{url}            
\usepackage{booktabs}       
\usepackage{amsfonts}       
\usepackage{nicefrac}       
\usepackage{microtype}      
\usepackage{xcolor}         
\usepackage{graphicx}
\usepackage{mathtools}
\usepackage{amsmath}
\usepackage{amssymb}
\usepackage{color, colortbl}
\usepackage{cancel}
\usepackage{caption}
\usepackage{subcaption}
\usepackage{algpseudocode}
\usepackage{algorithm}
\usepackage{multirow}
\usepackage{xfrac}
\usepackage{lineno}
\usepackage{algpseudocode}
\usepackage{algorithm}
\usepackage{array}
\usepackage{fixltx2e}
\usepackage{stfloats}
\usepackage{picins}
\usepackage{svg}

\usepackage[font=footnotesize,labelfont=bf]{caption}
\usepackage[font=footnotesize,labelfont=bf]{subcaption}

\journal{Neurocomputing}

\begin{document}

\begin{frontmatter}

\title{Disentanglement with Factor Quantized Variational Autoencoders}


\author[1,2]{Gulcin Baykal\corref{cor1}}
\cortext[cor1]{Corresponding author}
\ead{baykalg@imada.sdu.dk}

\author[1]{Melih Kandemir}
\ead{kandemir@imada.sdu.dk}

\author[3]{Gozde Unal}
\ead{gozde.unal@itu.edu.tr}

\affiliation[1]{organization={Department of Mathematics and Computer Science,
                              University of Southern Denmark},
                city={Odense},
                country={Denmark}}

\affiliation[2]{organization={Department of Computer Engineering, 
                              Istanbul Technical University},
                city={Istanbul},
                country={Türkiye}}
                
\affiliation[3]{organization={Department of AI and Data Engineering, 
                              Istanbul Technical University},
                city={Istanbul},
                country={Türkiye}}

\begin{abstract}

Disentangled representation learning aims to represent the underlying generative factors of a dataset in a latent representation independently of one another. In our work, we propose a discrete variational autoencoder (VAE) based model where the ground truth information about the generative factors are not provided to the model. We demonstrate the advantages of learning discrete representations over learning continuous representations in facilitating disentanglement. Furthermore, we propose incorporating an inductive bias into the model to further enhance disentanglement. Precisely, we propose scalar quantization of the latent variables in a latent representation with scalar values from a global codebook, and we add a total correlation term to the optimization as an inductive bias. Our method called FactorQVAE combines optimization based disentanglement approaches with discrete representation learning, and it outperforms the former disentanglement methods in terms of two disentanglement metrics (DCI and InfoMEC) while improving the reconstruction performance. Our code can be found at \url{https://github.com/ituvisionlab/FactorQVAE}.

\end{abstract}

\begin{keyword}
Disentanglement, Discrete Representation Learning, Vector Quantized Variational Autoencoders
\end{keyword}

\end{frontmatter}


\section{Introduction}\label{sec:introduction}

Discovering and extracting meaningful representations from raw data relevant to the task at hand has been the main purpose of representation learning. While the answer to the fundamental question of \textit{what makes a representation good} varies based on the task to perform, one possible answer is being "disentangled" \citep{bengio2013rl}. A disentangled representation depicts the distinct and independent underlying generative factors of the data in its different, interpretable components. As an example from the image modality, a disentangled representation captures the underlying generative factors such as object color, shape, orientation, or lighting, with its separate dimensions. 

Disentangled representations advance performance in numerous unlike domains and tasks such as image retrieval \citep{dai2022feature}, fairness in representations \citep{jang2024fair}, social recommendation \citep{wang2025causal}, and style transfer \citep{hu2023token}. Despite its contributions to various domains, unsupervised disentanglement has proven to be challenging, and even impossible, without certain restrictive assumptions \citep{locatello2019challenging}. Disentangled representation learning can be considered as an ill-posed problem as there is no unique way to disentangle the generative factors in data \citep{horan2021when}. Lack of knowledge about the true generative factors for most of the datasets makes it more difficult to obtain and evaluate disentangled representations. Therefore, the success of disentangled representation learning heavily relies on the inductive biases deployed into the model. Disentanglement may be induced via architecture design \citep{leeb2023structure}, restrictive constraints used as regularizers \citep{whittington2023disentanglement}, and enhanced loss functions \citep{chen2018isolating}.

A recent work proposes QLAE \citep{hsu2023disentanglement}, a VAE based model that uses discrete representations as a better inductive bias for disentanglement. Discrete representation learning has gained further importance with the tremendous success of vector quantization for representation learning with VQVAE \citep{oord2017neural}, and has been used in recent large generative models like Latent Diffusion Models \citep{rombach2021highresolution} and DALL-E \citep{ramesh2021zeroshot}. While discrete representations are better suited for expressing categories that shape the observation space \citep{oord2017neural}, \citet{hsu2023disentanglement} use this rationale for disentanglement, and quantize the continuous representations with a small set of scalar values in order to force the model to assign constant meaning to each value. For each dimension of the representation, \citet{hsu2023disentanglement} define a separate set of scalar values, referred to as a codebook, and quantize each value using a scalar from its corresponding codebook. The quantizing value is selected based on its proximity to the corresponding variable in the continuous representation. Although \citet{hsu2023disentanglement} demonstrate that using individual codebooks for each dimension, rather than a single global codebook, improves disentanglement, we identify certain challenges with this approach. 

In QLAE, the quantization of each latent variable is restricted to its own codebook that includes a predefined number of scalar values, \textbf{k}. Assume two generative factors, \textit{shape} and \textit{orientation} that are ideally captured with different latent variables. These factors have a different number of options, \textbf{n} and \textbf{m} where $\textbf{n} < \textbf{m}$, respectively. Therefore, the range of values in the latent representation may vary to capture these options, such that the range of latent values for \textit{orientation} might be wider since $\textbf{n} < \textbf{m}$. Quantizing the latent values of different generative factors using individual codebooks can be limiting in terms of representation capacity. For example, using \textbf{k} scalars may be insufficient to represent \textbf{m} possible \textit{orientation}s, while at the same time, \textbf{k}  could be excessive for representing \textbf{n} possible \textit{shape}s. Even if the codebook size \textbf{k} is large enough to cover the generative factor with the most options, this would lead to redundancy in the codebook for other generative factors with fewer options, as the codebook size remains equal for each factor. 

Although QLAE’s use of individual codebooks promotes disentanglement, its fixed-size, per-dimension quantization introduces inefficiencies and limits representational flexibility, especially when generative factors have diverse cardinalities. To overcome these limitations, we propose FactorQVAE, a novel discrete latent variable framework explicitly designed for disentangled representation learning. Unlike prior discrete models such as VQVAE and dVAE, which primarily enhance reconstruction quality or sampling efficiency, FactorQVAE integrates scalar quantization using a single global codebook with a total correlation regularizer. As discussed in Section~\ref{sec:related_work}, total correlation is a prominent regularizer for disentanglement, encouraging statistical independence among latent variables by adding a specific penalty term to the loss function \citep{kim2018disentangling}. Integrating scalar quantization with total correlation is challenging because discrete latents must remain differentiable and limited codebook capacity must be managed, which FactorQVAE addresses via a stochastic categorical posterior and a global codebook. These innovations translate directly into practical benefits for real-world tasks like robotics, visual reasoning, and fairness-aware systems. For instance, disentangling object attributes facilitates robotic adaptability, while isolating sensitive factors enhances transparency in fairness-critical systems. Thus, FactorQVAE bridges fundamental theoretical advances with meaningful practical applications.

We summarize our contributions as follows:
\begin{itemize}
    \item We introduce FactorQVAE that originally combines factorization as a regularizer and the discrete VAEs for disentanglement, using scalar quantization with a single global codebook to overcome the representational inefficiencies of fixed per-dimension codebooks.
    \item We report the effects of factorization and discretization on disentanglement individually and demonstrate the effectiveness of their novel combination through a stochastic categorical posterior for improved disentangled representations. 
    \item We redesign the training frameworks of two discrete VAE models, VQVAE \citep{oord2017neural} and dVAE \citep{ramesh2021zeroshot}, to enhance disentanglement and systematically evaluate the trade-offs between scalar and vector quantization approaches. 
\end{itemize}
In our work, we compare FactorQVAE with recent and state-of-the-art disentanglement models using three different datasets and two different disentanglement evaluation metrics. We demonstrate that our model performs close to or better than the other methods in terms of disentanglement.

\section{Related Work}\label{sec:related_work}

\textbf{Disentangled Representation Learning: } Methods for disentangled representation learning can be categorized by their model types, supervision levels, and independence assumptions \citep{wang2024disentangled}. We choose the VAE as the base model architecture over alternatives such as generative adversarial networks (GAN) \citep{goodfellow2014generative} and diffusion models \citep{ho2020denoising}. 

Several VAE-based disentanglement approaches have been proposed, each with limitations motivating our work. \citet{higgins2017beta} introduce $\beta$-VAE, where a higher $\beta$ creates a fundamental reconstruction–disentanglement trade-off, significantly reducing reconstruction fidelity and producing uninformative latent representations. \citet{kim2018disentangling}'s FactorVAE partially mitigates these issues by explicitly penalizing total correlation, yet the hyperparameter-sensitive trade-off persists. \citet{chen2018isolating} propose decomposing the VAE loss with multiple separate hyperparameters, which enhances flexibility but exacerbates hyperparameter tuning challenges. \citet{whittington2023disentanglement}'s BioAE incorporates biologically inspired constraints, which introduce domain-specific assumptions limiting its generalizability. QLAE proposed by \citet{hsu2023disentanglement}, another discrete latent representation approach, requires unusually strong regularization (e.g., high weight decay), and its per-dimension codebooks restrict latent expressiveness.

Recent methods also face distinct practical challenges. \citet{meo2024alpha}'s $\alpha$-TCVAE combines variational and conditional entropy bottlenecks but remains sensitive to the $\alpha$ hyperparameter balancing diversity and reconstruction quality. \citet{li2025revisit}'s PDisVAE leverages partial correlation for group-wise independence but struggles with complex hyperparameter optimization and selecting optimal latent groupings. \citet{uscidda2025disentangled}'s approach based on the Gromov-Monge Gap regularizer using optimal transport theory suffers from quadratic computational complexity and demands careful hyperparameter tuning.

In terms of supervision, since unsupervised disentanglement is theoretically impossible without inductive biases \citep{locatello2019challenging}, weakly-supervised \citep{locatello2020weakly} and self-supervised \citep{eastwood2023selfsupervised} methods have also been proposed for disentanglement. While all the aforementioned methods generally assume statistical independence between the generative factors and the variables in the latent representation, some approaches also consider causal relationships between the generative factors \citep{yang2021causalvae}.

In contrast, our proposed FactorQVAE is an unsupervised VAE-based method explicitly assuming independent generative factors and uniquely integrating discrete latent quantization with total correlation regularization. Unlike prior methods, FactorQVAE simultaneously optimizes stability, reconstruction quality, and disentanglement, significantly mitigating hyperparameter sensitivity. This translates directly into practical advantages, including more robust performance across diverse datasets and easier training procedures.

Note that generative models such as Latent Diffusion Models \citep{rombach2021highresolution}, which implicitly leverage disentanglement through multimodal inputs for high-quality image generation, fall outside our scope, as our focus remains explicitly on structured disentangled representations. Likewise, recent approaches \citep{liu2024improving, ozcan2025generalized} that prioritize disentanglement for only selected latent dimensions while allowing others to remain entangled address a different aspect of the problem. In contrast, FactorQVAE relies on discrete quantization and treats all latent dimensions uniformly, ensuring balanced disentanglement across the representation. In addition, recent work has explored disentanglement in broader domains \citep{pu2025dear, cai2025learning}, reflecting a growing interest in adapting disentangled representation learning beyond 3D-structured benchmarks. Our additions position FactorQVAE within this broader context while highlighting its distinct contribution in discrete, factorized representation learning.

\textbf{Discrete Representation Learning: } After \citet{oord2017neural} demonstrated the significant advantages of discrete representation learning, vector quantization, introduced as a discretization method in \citep{oord2017neural}, gained popularity for various representation learning tasks across different modalities, including image generation \citep{esser2020taming}, music generation \citep{dhariwal2020jukebox}, and text decoding \citep{kaiser18fast}. 

Vector quantization is also utilized in \citep{wu2024neural} for object-centric representation learning using multiple codebooks, where each codebook captures a distinct semantic meaning. An object's representation is then obtained by combining embeddings from these semantic codebooks. The study demonstrates that vector quantization with multiple codebooks aids in disentangling semantic features in object-centric representation learning. In contrast to \citep{wu2024neural}, which relies on the known number of ground truth semantic factors to determine the number of codebooks, our method is unsupervised and specifically designed for disentanglement.

\citet{mercatali2021disentangling} propose to use discrete VAEs for disentangling the generative factors in natural language. Our method differs from \citep{mercatali2021disentangling}, as their approach requires knowledge of the number of possible values for each generative factor during training, whereas our method is fully unsupervised.

\citet{hsu2024tripod} also highlight the importance of discrete representations for disentanglement, along with other inductive biases guiding the encoder and decoder for disentanglement. Our method differs from \citep{hsu2024tripod} in terms of the discretization method, as we employ codebook learning, and in the specific inductive biases added to promote disentanglement.
\section{Background}\label{sec:background}

\subsection{Discrete Variational Autoencoders}\label{sec:background_dvae}

Discrete VAEs aim to represent high-dimensional data $x$ using a low dimensional, discrete latent representation $z$ by maximizing the Evidence Lower Bound (ELBO) objective:
\begin{equation}
    \mathcal{L}_{\text{ELBO}} = \mathbb{E}_{q(z|x)}[\log p(x|z)] - \text{KL}[q(z|x) \| p(z)],
\label{eq:elbo}
\end{equation}
where the generative model $p(x|z)$ is implemented as a decoder, the approximated posterior $q(z|x)$ is implemented as an encoder, and $p(z)$ is the prior. This objective addresses the challenge of approximating the posterior distribution $p(z|x)$, which is intractable due to the complexity of computing $p(x)$. Variational inference is used to approximate this posterior with a tractable distribution $q(z|x)$ and optimize the ELBO, making $q(z|x)$ as close as possible to the true posterior.

In discrete VAEs, the $d$-dimensional latent variables $z$ are sampled from a Categorical distribution as $z \sim Cat(z|\pi)$. The encoder outputs unnormalized log probabilities $l=[l_1, \ldots, l_d]$, and the probability masses $\pi$ of this Categorical distribution are obtained as $\text{softmax}(l)$. Subsequently, $z$ are fed into the decoder to reconstruct $x$.

VQVAE \citep{oord2017neural} is a discrete VAE variant featuring a learnable codebook $\mathcal{M} \in \mathbb{R}^{K \times C}$ composed of $K$ number of $C$-dimensional embeddings, which is trained to represent a dataset. In VQVAE, an observation $x$ is represented by the selected embeddings from $\mathcal{M}$. At first, the continuous latent representation $z_e(x) = \mathcal{E}_\theta (x)$ where $z_e(x) \in \mathbb{R}^{N \times N \times C}$ is obtained by the encoder $\mathcal{E}_\theta$. Then, the Euclidean distances between $z_e(x)$ and the embeddings in $\mathcal{M}$ are calculated. The discrete latent variables $z \in \mathbb{R}^{N \times N \times K}$ are sampled as one-hot from the posterior $q(z|x)$. The deterministic posterior Categorical distribution is defined as follows:
\begin{equation}
    q(z=k|x) =
        \begin{cases}
            1  & \quad \text{if } k=\text{argmin}_{j}\|z_e(x)_i-e_j\|_2 \\
            0  & \quad \text{otherwise} 
        \end{cases}
    \label{eq:deterministic_posterior}
\end{equation} 
where $k,j \in \{1, \dots, K\}$, $i \in {1, \ldots, N \times N}$, $z_e(x)_i$ is the $i^{th}$ embedding in $z_e(x)$, and $e_j$ is the $j^{th}$ codebook embedding. The discrete latent variables $z$ are used to retrieve the corresponding embeddings from $\mathcal{M}$ obtaining  the quantized representation $z_q(x)$ through matrix multiplication, $z_q(x)=z*\mathcal{M}$. Finally, $z_q(x) \in \mathbb{R}^{N \times N \times C}$ is fed into the decoder $\mathcal{D}_\phi$, and $x = \mathcal{D}_\phi(z_q(x))$ is obtained.

While VQVAE's encoder learns a continuous representation to be directly quantized with the codebook $\mathcal{M}$ based on the distances, another discrete VAE variant dVAE \citep{ramesh2021zeroshot}'s encoder outputs $z_e(x) \in \mathbb{R}^{N \times N \times K}$ that is treated as the unnormalized log probabilities $l$ of a Categorical distribution over $K$ number of embeddings in $\mathcal{M}$, and $z \sim Cat(z|\text{softmax}(l))$ is attained. Instead of feeding the discrete latent variables $z$ to the decoder directly, they are used to retrieve the corresponding codebook embeddings to construct $z_q(x)$ as in VQVAE.

In our work, we investigate the performance of both dVAE and VQVAE in terms of disentanglement and incorporate our method into both to further enhance disentanglement.

\subsection{Disentangled Representation Learning}\label{sec:background_drl}

Assume that the observation $x$ sampled from the generative model $p(x|z)$ has the underlying generative factors $\textbf{s} = (s_1, \ldots, s_F)$, and the factorized density $p(z) = \prod_{i=1}^{d}p(z_i)$ holds for $z$. Disentanglement is the process of learning to represent $F$ number of mutually independent underlying generative factors $\textbf{s}$ where $p(\textbf{s}) = \prod_{j=1}^{F}p(s_j)$ holds, in the separate, independent $d$ number of components of the latent representation $z$. Single latent variables should capture changes in a specific underlying factor, while remaining relatively unaffected by changes in other factors \citep{bengio2013rl}.

In our work, we seek for a model that can achieve $\mathcal{D}_\phi(\mathcal{E}_\theta(x)) \sim x$ with a high degree of disentanglement.

\subsection{Disentanglement Metrics}\label{sec:background_metrics}

Disentanglement metrics can be categorized as \textit{intervention-based}, \textit{predictor-based}, and \textit{information-based} \citep{carbonneau2024measuring}. Intervention-based metrics evaluate disentanglement by fixing factors, creating subsets of data points, and comparing the corresponding codes and factors within these subsets to produce a score. Predictor-based metrics involve training a regressor/classifier $f$ to predict generative factor realizations from latents $f(z) \mapsto s$, followed by analyzing the predictor to assess the usefulness of each latent dimension in accurate predicting the generative factors. Information-based metrics calculate a disentanglement score by estimating the mutual information between the latents and the generative factors.

In our work, we select "Disentanglement, Completeness and Informativeness (DCI)" metric \citep{eastwood2018dci} from the predictor-based metrics as they are generally the best performing solutions \citep{carbonneau2024measuring}, and "Modularity, Explicitness, and Compactness (InfoMEC)" from the information-based metrics as it is most up-to-date metric proposed in the literature \citep{hsu2023disentanglement} for evaluation.

\subsubsection{DCI}

To calculate DCI scores, $F$ number of regressors or classifiers are trained to predict generative factors $\textbf{s} = (s_1, \ldots, s_F)$ from the latent variables $z$. Assume $f_j(z \mapsto s_j)$ is a predictor mapping $z$ to $j^{th}$ generative factor $s_j$. An importance matrix $M \in \mathbb{R}^{d \times F}$ can be formed such that $M_{ij}$ is the magnitude of the weight learned by the predictor $f_j$ which indicates the relationship between the $i^{th}$ latent variable $z_i$ and $s_j$. Here $d$ refers to the number of latent variables.

Disentanglement ($\mathsf{D}$), also known as modularity, refers to the extent to which $z$ separates the generative factors $\textbf{s}$, with each latent variable capturing at most one distinct generative factor. $D$ is calculated as follows:
\begin{align*}
    P_{ij} &= M_{ij} / \textstyle\sum_{k=0}^{F-1}M_{ik}, \\
    \mathsf{D}_i &= 1 + \textstyle\sum_{k=0}^{F-1}P_{ik}\text{log}_{F}P_{ik}, \\
    \rho_i &= \textstyle\sum_j M_{ij} / \textstyle\sum_{ij}M_{ij}, \\
    \mathsf{D} &= \textstyle\sum_i \rho_i\mathsf{D}_i,
\end{align*}
where $P_{ij}$ represents $z_i$'s probability of being important for predicting each $s_j$, $\mathsf{D}_i$ is the disentanglement score for $z_i$, and $\rho_i$ is the weight of $\mathsf{D}_i$ for average disentanglement score $\mathsf{D}$.

Completeness ($\mathsf{C}$), also known as compactness, refers to the extent to which each generative factor $s_j$ is captured by a single latent variable. $\mathsf{C}$ is calculated as follows:
\begin{align*}
    \tilde{P}_{ij} &= M_{ij} / \textstyle\sum_{k=0}^{d-1}M_{kj}, \\
    \mathsf{C}_j &= 1 + \textstyle\sum_{k=0}^{d-1}\tilde{P}_{kj}\text{log}_{d}\tilde{P}_{kj}, \\
    \rho_j &= \textstyle\sum_i M_{ij} / \textstyle\sum_{ij}M_{ij}, \\
    \mathsf{C} &= \textstyle\sum_j \rho_j\mathsf{C}_j,
\end{align*}
where $\tilde{P}_{ij}$ represents each $z_i$'s probability of being important for predicting $s_j$, $\mathsf{C}_i$ is the completeness score for $s_j$, and $\rho_j$ is the weight of $\mathsf{C}_j$ for average completeness score $\mathsf{C}$.

Informativeness ($\mathsf{I}$), also known as explicitness, is the extent to which $z$ captures information about the underlying generative factors $\textbf{s}$. $\mathsf{I}$ is calculated as the mean accuracies of $F$ different predictors that are essentially learning the relationship between $z$ and $\textbf{s}$.

\subsubsection{InfoMEC}

Instead of training predictors and using their weights and accuracies to evaluate disentanglement, mutual information (MI) between $z_i$ and $s_j$ can be used as a notion of informativeness. MI quantifies the amount of information shared between the latent variable $z_i$ and the underlying factor $s_j$, providing a clear and objective metric for how well $z_i$ captures $s_j$. By focusing on MI, we can assess disentanglement in terms of how effectively each latent variable encodes distinct factors of variation, without the need for potentially biased or indirect measures from predictor performance. MI between $z_i$ and $s_j$ can be defined as:
\begin{align*}
    I(z_i;s_j) &= D_{\text{KL}}\left(p(s_j,z_i) \| p(s_j)p(z_i)\right) \\
    &= H(s_j) - H(s_j | z_i),
\end{align*}
where $H(.)$ is the entropy function. Normalized mutual information (NMI) which depicts the relationship between $z$ and $\textbf{s}$ as a matrix can be defined as:
\begin{align*}
    \text{NMI}(z_i, s_j) &= \frac{I(z_i, s_j)}{H(s_j)}.
\end{align*}
To compute the modularity of $z_i$, \citet{chen2018isolating} propose to use the gap between the two largest elements in the $i^{th}$ row of NMI, while \citet{hsu2023disentanglement} prefer using the ratio of the largest element in the $i^{th}$ row to the row sum, and propose InfoM as the average modularity which is defined as:
\begin{align*}
    \text{InfoM} = \left(
        \frac{1}{d} \sum_{i=1}^{d} \frac{\text{max}_{j} \text{NMI}_{ij}}{\sum_{j=1}^{F} \text{NMI}_{ij}} - \frac{1}{F}
    \right)\bigg/\left(
        1 - \frac{1}{F}
    \right).
\end{align*}
To compute the average compactness, \citet{hsu2023disentanglement} propose InfoC which is defined as:
\begin{align*}
    \text{InfoC} = \left(
        \frac{1}{F} \sum_{j=1}^{F} \frac{\text{max}_{i} \text{NMI}_{ij}}{\sum_{i=1}^{d} \text{NMI}_{ij}} - \frac{1}{d}
    \right)\bigg/\left(
        1 - \frac{1}{d}
    \right).
\end{align*}
To compute explicitness, \citet{hsu2023disentanglement} borrow the framework of predictive $\mathcal{V}$-information \citep{xu2020theory}, and follows these steps:
\begin{align*}
    H_{\mathcal{V}}(s_j | z) &= \inf_{f \in \mathcal{V}}\mathbb{E}_{s \sim p(s),z \sim p(z|s)}\left[ -\log p(s_j | f(z)) \right], \\    
    H_{\mathcal{V}}(s_j | \varnothing) &= \inf_{f \in \mathcal{V}}\mathbb{E}_{s \sim p(s)}\left[ -\log p(s_j | f(\varnothing)) \right], \\    
    I_{\mathcal{V}}(z \to s_j) &= H_{\mathcal{V}}(s_j | \varnothing) - H_{\mathcal{V}}(s_j | z), \\
    \text{NMI}_{\mathcal{V}}(z \to s_j) &= \frac{I_{\mathcal{V}}(z \to s_j)}{H_{\mathcal{V}}(s_j | \varnothing)}, \\
    \text{InfoE} &= \frac{1}{F}\sum_{j=1}^{F}\text{NMI}_{\mathcal{V}}(z \to s_j),
\end{align*}
where $\mathcal{V}$ is an allowable function class for the computation of information, $H_{\mathcal{V}}(s_j | z)$ is the predictive conditional $\mathcal{V}$-entropy, $H_{\mathcal{V}}(s_j | \varnothing)$ is the marginal $\mathcal{V}$-entropy, $I_{\mathcal{V}}(z \to s_j)$ is the predictive $\mathcal{V}$-information of $s_j$, $\text{NMI}_{\mathcal{V}}(z \to s_j)$ is the normalized predictive $\mathcal{V}$-information, and InfoE is the average explicitness.

In our work, we follow the literature and set $d > F$. Specifically, we set $d = 2*F$ where $F$ changes based on the dataset. As a consequence of this inequality, achieving both perfect disentanglement (modularity) and perfect completeness (compactness) is impossible. Therefore, achieving a better disentanglement (modularity) value is prioritized.

\section{Method}\label{sec:method}

\begin{figure}[t]
    \centering
    \includegraphics[width=\textwidth]{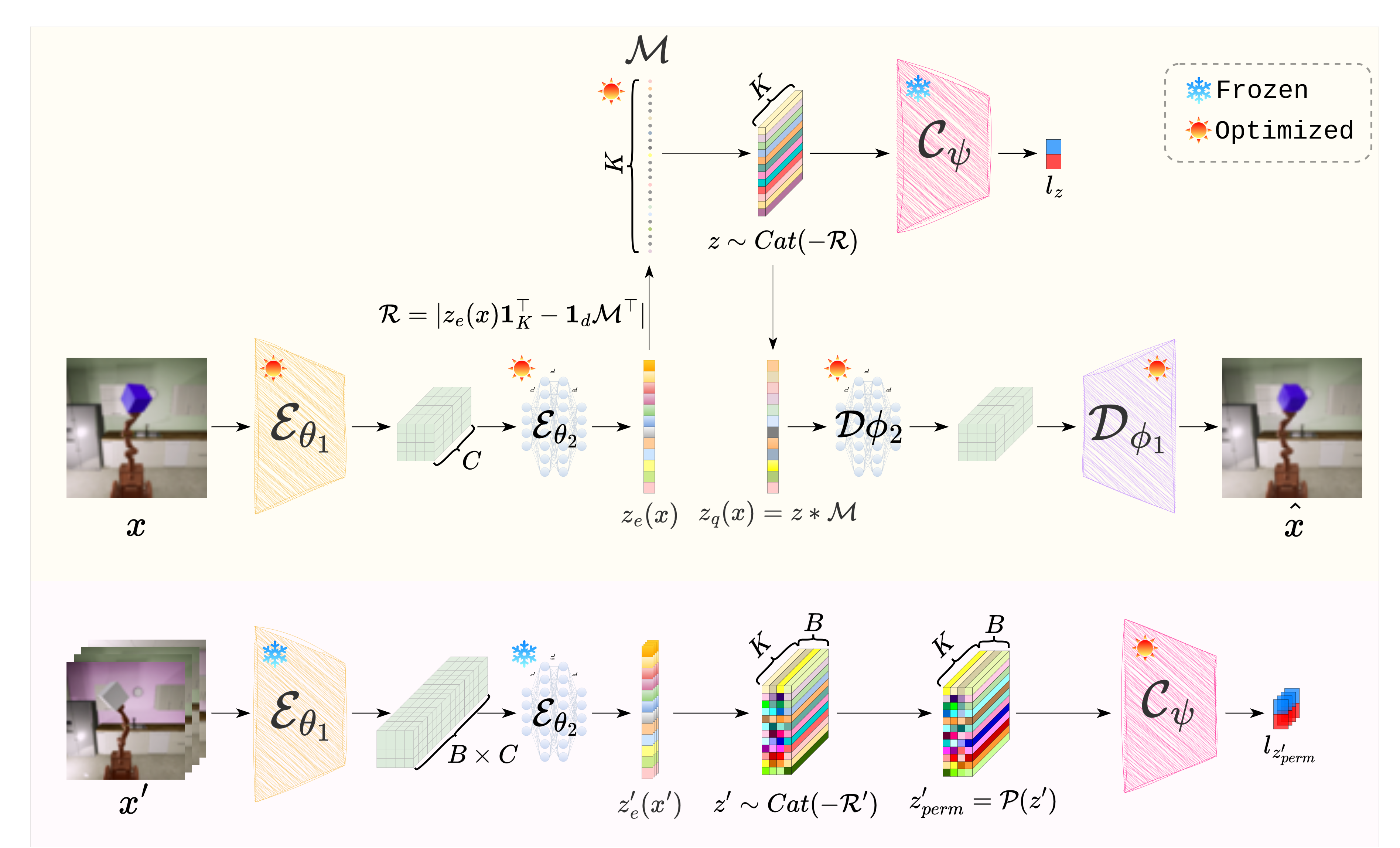}
    \caption{At the first stage (yellow background), an input $x$ is encoded into a latent representation $z_e(x)$ by the encoder $\mathcal{E}_{\theta_1}$, followed by some nonlinear operations $\mathcal{E}_{\theta_2}$. Each latent variable in $z_e(x)$ is quantized with the colored scalars from the codebook $\mathcal{M}$ whose indices $z$ are sampled based on the distances $\mathcal{R}$ between $z_e(x)$ and $\mathcal{M}$. The quantized latent representation $z_q(x)$ is transformed by nonlinear operations $\mathcal{D}_{\phi_2}$, and fed into the decoder $\mathcal{D}_{\phi_1}$ to reconstruct $x$. The discriminator $\mathcal{C}_\psi$ outputs log probabilities that its input is sampled from $q(z)$ rather than from $\bar{q}(z)$. At second stage (pink background), a new data batch $x'$ is sampled. Permuter $\mathcal{P}$ permutes the one-hot indices $z'$ across the latent dimensions, and yields $z'_{perm}$ (best viewed in PDF with zoom).}
    \label{fig:main_figure}
\end{figure}

The overview of our method FactorQVAE is shown in Figure~\ref{fig:main_figure}. In FactorQVAE, we originally integrate discrete representation learning with factorization for enhanced disentanglement. While we use a single sample in the first stage to explain discrete representation learning (Figure~\ref{fig:main_figure}-yellow highlighted background), we prefer batch view to explain factorization better (Figure~\ref{fig:main_figure}-pink highlighted background).

We start by making substantial adjustments to the vector quantization process described in Section~\ref{sec:background_dvae} in order to tailor it for disentanglement. Referring to Figure~\ref{fig:main_figure}, we view the output of $\mathcal{E}_{\theta_1}$ as a feature tensor depicting spatial information about input $x$, and further transform the features by  nonlinear operations $\mathcal{E}_{\theta_2}$ to obtain a vector latent representation $z_e(x) \in \mathbb{R} ^ {d*C \times 1}$. Hence, $z_e(x)$ now captures the underlying generative factors with its latent variables, rather than spatial information. 

Originally, VQVAE quantizes the $C$ dimensional vectors in $z_e(x)$ with the embeddings in the codebook $\mathcal{M}$ with a deterministic choice based on the distances. Instead of vector quantization, we propose scalar latent variable quantization with a single, global codebook $\mathcal{M} \in \mathbb{R}^{K \times 1}$ consisting $K$ number of scalar values, and name this model QVAE, rather than VQVAE. Consequently, the output of $\mathcal{E}_{\theta_1}$ which is an $N \times N \times C$ dimensional feature tensor is transformed into $z_e(x) \in \mathbb{R}^{d \times 1}$ so that each latent variable in $z_e(x)$ can be quantized with a scalar from $\mathcal{M}$. 

In designing our framework, we intentionally employ scalar quantization instead of alternative approaches such as vector quantization. Scalar quantization explicitly constrains the latent representation’s information capacity, promoting an effective balance between reconstruction and disentanglement. In contrast, vector quantization provides higher-dimensional latent embeddings with greater representational capacity, which can inadvertently lead the model to prioritize reconstruction accuracy over disentangled representations by encoding redundant or correlated information. From an information-theoretic perspective, this difference can be understood as follows: vector quantization allows each codebook vector to capture correlations across multiple latent dimensions, thereby increasing inter-dimensional mutual information and making disentanglement more difficult. Scalar quantization, by contrast, enforces independent per-dimension assignments, which act as a stronger information bottleneck. This encourages alignment between individual latent variables and generative factors while reducing redundancy between dimensions. Hence, our deliberate choice of scalar quantization provides both an intuitive and theoretically motivated inductive bias toward disentanglement. We validate this decision empirically in Section~\ref{sec:experiments} and Table~\ref{tab:codebook_design}.

Apart from the representation design, we further modify the variational family. Rather than using a deterministic categorical posterior as in Equation~\ref{eq:deterministic_posterior}, we define a stochastic categorical posterior proposed by \citet{sonderby2017continuous} as:
\begin{align}
    \mathcal{R} &= |z_e(x)\mathbf{1}_K^\top - \mathbf{1}_d\mathcal{M}^\top|, \label{eq:distance}\\
    q(z|x) &= Cat(-\mathcal{R}). \label{eq:stochastic_posterior}
\end{align}
In Equation~\ref{eq:distance}, we calculate the distance matrix $\mathcal{R} \in \mathbb{R}^{d \times K}$ between $z_e(x) \in \mathbb{R}^{d \times 1}$ and $\mathcal{M}$. We use $\mathbf{1}_K$ and $\mathbf{1}_d$ vectors of ones with $K$ and $d$ dimensions, respectively, for computational ease in calculation of the distance matrix $\mathcal{R}$. $\mathcal{R}$ holds $i^{th}$ latent variable's distances to $\mathcal{M}$ in its $i^{th}$ row.

Since the closest scalar in $\mathcal{M}$ has the lowest distance, we use $-\mathcal{R}$ as the parameters of the Categorical distribution to be able to sample the closest scalar with a higher probability in Equation~\ref{eq:stochastic_posterior}. Sampling from the stochastic posterior $q(z|x)$ yields the discrete variables $z \in \mathbb{R}^{d \times K}$ which are essentially the one-hot representations of the indices of the quantizing scalars for $d$ number of latent variables. Sampling from a Categorical distribution is inherently non-differentiable, which complicates end-to-end training when combined with the total correlation regularizer. To overcome this, we adopt the Gumbel–Softmax distribution as a differentiable approximation \citep{maddison2016concrete, jang2016categorical}, allowing gradient-based optimization while preserving the discrete nature of the latent variables. These differentiable samples will be essential when we add the total correlation term to the optimization. 

The quantized representation $z_q(x) \in \mathbb{R}^{d \times 1}$ is obtained by a matrix multiplication of the discrete variables $z$ and $\mathcal{M}$. $z_q(x)$ is originally fed into the decoder $\mathcal{D}_{\phi_1}$ in VQVAE as it is the quantization of the feature tensor. In our case, we further transform the latent vector $z_q(x)$ into a feature tensor by nonlinear layers in $\mathcal{D}_{\phi_2}$ before feeding it into $\mathcal{D}_{\phi_1}$ to reconstruct input $x$.

Although FactorQVAE is first built upon a VQVAE based method with major modifications, the dVAE framework can be also used as the discrete representation learning part of our model. While $\mathcal{R}$ in Equation~\ref{eq:distance} is calculated to define the parameters of the Categorical distribution in FactorQVAE, we can directly learn these parameters by $\mathcal{E}_\theta$ as in dVAE. Based on dVAE, we can learn $z_e(x) \in \mathbb{R}^{d \times K}$ which essentially represents the parameters of the Categorical distribution over each codebook element for each latent dimension, and perform $z \sim Cat(z_e(x))$. We explore the importance of the design of the discrete representation learning module by modeling FactordVAE based on dVAE, along with FactorQVAE.

We have covered the discrete representation learning aspect of the framework in the first stage. Next, we explain how we add factorization to the training as a regularizer that enforces disentanglement. As suggested by \citet{kim2018disentangling}, a new term called "total correlation" can be incorporated into the ELBO, which is then maximized over the overall loss:
\begin{align}
    \mathcal{L}(\mathcal{M}, \theta_1, \theta_2, \phi_1, \phi_2) = &\frac{1}{N}\sum_{i=1}^N\left[ \mathbb{E}_{q(z|x^i)}\left[\log p(x^i|z)\right]-\beta\text{KL}(q(z|x^i)\|p(z))\right] \nonumber \\
    & -\gamma \text{KL}(q(z)\|\bar{q}(z)),
    \label{eq:loss}
\end{align}
where the total correlation term is given by $\text{KL}(q(z)\|\bar{q}(z))$. Here, $q(z) = \frac{1}{N}\sum_{i=1}^Nq(z|x^i)$ is the marginal posterior over the entire dataset, and $\bar{q}(z) = \prod_{j=1}^dq(z_j)$ indicates factorial distribution. The total correlation term encourages the marginal posterior to factorize, thereby promoting disentanglement. 

Since the total correlation calculations are intractable, a common approach is to resort to a practical solution to approximate the total correlation \citep{kim2018disentangling}. First, sampling from $q(z)$ is performed by sampling from $q(z|x^i_B)$ where $x^i_B$ is a sample within the randomly selected batch $B$. For approximating $\bar{q}(z)$, we can sample a new batch $B'$ and sample $z' \sim q(z|x^i_{B'})$, then randomly permute across the batch for each latent dimension by the permutation operator $\mathcal{P}$ to obtain $z'_{perm}=\mathcal{P}(z')$. This permutation operation is visually exemplified in Figure~\ref{fig:main_figure} in the second stage. It is important to note that in FactorQVAE, each latent dimension consists of $K$ dimensional one-hot vectors unlike FactorVAE \citep{kim2018disentangling} where each latent dimension is one dimensional. 

Essentially, by permuting the samples across each latent dimension independently, we create a new set of samples that behave as if each dimension $z_j$ was independently sampled from $q(z_j)$. This independence and matching distribution are exactly what $\bar{q}(z)$ represents, and thus the distribution of the permuted batch $z'_{perm}$ approximates $\bar{q}(z)$ when the batch size is sufficiently large.

We minimize the KL divergence between $q(z)$ and $\bar{q}(z)$ using the density ratio trick:
\begin{align}
    \text{KL}(q(z)\|\bar{q}(z)) &= \mathbb{E}_{q(z)}\left[ \log \frac{q(z)}{\bar{q}(z)}\right], \label{eq:density_ratio} \\ 
    &\approx \mathbb{E}_{q(z)}\left[\log \frac{\mathcal{C}_\psi(z)}{1-\mathcal{C}_\psi(z)}\right] \label{eq:density_ratio_approx},
\end{align}
where $\mathcal{C}_\psi$ is a discriminator that approximates the density ratio in Equation~\ref{eq:density_ratio}. Assume $\mathcal{C}_\psi(z)$ is the estimated probability that $z$ are sampled from $q(z)$ rather than $\bar{q}(z)$. In order to train such a discriminator that can differentiate between the samples from $q(z)$ and $\bar{q}(z)$, $\mathcal{C}_\psi$ should be optimized by minimizing the following loss function:
\begin{equation}
    \mathcal{L}(\psi) = \mathbb{E}_{z \sim q(z)}\left[\log (\mathcal{C}_\psi(z))\right]+\mathbb{E}_{z'_{perm} \sim \bar{q}(z)}\left[\log (1-\mathcal{C}_\psi(z'_{perm}))\right]. \label{eq:disc_update}
\end{equation}
The second stage in Figure~\ref{fig:main_figure} shows the training procedure of $\mathcal{C}_\psi$. As we need to backpropagate the gradients from $\mathcal{C}_\psi$ to the other modules, $z$ must remain differentiable, which is ensured by our modification of variational family compared to VQVAE. Algorithm~\ref{alg:train} presents the pseudocode of the overall FactorQVAE training that we propose.

\begin{algorithm}[t!]
\caption{Training algorithm of FactorQVAE}\label{alg:train}
\begin{algorithmic}
\State {\bfseries Input:} Observations $(\textbf{x}_{\text{train}}^i)_{i=1}^N$, batch size $B$, $\beta$, $\gamma$, optimizers $g_1$ and $g_2$ 
\State Initialize the network parameters $\theta^{[0]}_1, \theta^{[0]}_2, \phi^{[0]}_1, \phi^{[0]}_2, \psi^{[0]}$, the codebook $\mathcal{M}^{[0]}$, 
\State temperature parameter $\tau^{[0]}$, learning rates $\alpha^{[0]}_1$ for $g_1$ and $\alpha^{[0]}_2$ for $g_2$.
\For{$t=1,2,\ldots,T$}
\State $x \leftarrow$ Random minibatch from $\textbf{x}_{\text{train}}$ of size $B$
\State $z_e(x) \leftarrow \mathcal{E}_{\theta^{[t-1]}_2}\left(\mathcal{E}_{\theta^{[t-1]}_1}(x)\right)$
\State $\mathcal{R} \leftarrow |z_e(x)\mathbf{1}_K^\top - \mathbf{1}_d\mathcal{M}^{[t-1]\top}|$
\State $z \sim \texttt{RelaxedOneHotCategorical}(\text{temperature}=\tau^{[t-1]}, \text{logits}=-\mathcal{R})$
\State $\mathcal{L}_1 \leftarrow \frac{1}{B}\sum_{i \in B}\left[-\log p(x^i|z^i)+\beta\text{KL}(q(z^i|x^i) \| p(z)))+\gamma\log \frac{\mathcal{C}_{\psi^{[t-1]}}(z^i)}{1-\mathcal{C}_{\psi^{[t-1]}}(z^i)}\right]$
\State $\mathcal{M}^{[t]}, \theta_1^{[t]}, \theta_2^{[t]}, \phi_1^{[t]}, \phi_2^{[t]}  \leftarrow g_1(\nabla_{\mathcal{M}, \theta_1, \theta_2, \phi_1, \phi_2}\mathcal{L}_1, \text{learning rate}=\alpha^{[t-1]}_1)$
\State $x' \leftarrow$ Random minibatch from $\textbf{x}_{\text{train}}$ of size $B$
\State $z'_e(x') \leftarrow \mathcal{E}_{\theta^{[t]}_2}\left(\mathcal{E}_{\theta^{[t]}_1}(x)\right)$
\State $\mathcal{R'} \leftarrow |z'_e(x')\mathbf{1}_K^\top - \mathbf{1}_d\mathcal{M}^{[t]\top}|$
\State $z' \sim \texttt{RelaxedOneHotCategorical}(\text{temperature}=\tau^{[t-1]}, \text{logits}=-\mathcal{R'})$
\State $z'_{perm} \leftarrow \mathcal{P}(z')$
\State $\mathcal{L}_2 \leftarrow \frac{1}{2B}\sum_{i \in B}\left[\log\mathcal{C}_{\psi^{[t-1]}}(z^i) + \log(1-\mathcal{C}_{\psi^{[t-1]}}({z'_{perm}}^i))\right]$
\State $\psi^{[t]} \leftarrow g_2(\nabla_{\psi}\mathcal{L}_2, \text{learning rate}=\alpha^{[t-1]}_2)$
\State $\tau^{[t]} \leftarrow \texttt{CosineAnneal}(\tau^{[t-1]}, t)$
\State $\alpha^{[t]}_1 \leftarrow \texttt{CosineAnneal}(\alpha^{[t-1]}_1, t)$
\State $\alpha^{[t]}_2 \leftarrow \texttt{CosineAnneal}(\alpha^{[t-1]}_2, t)$
\EndFor
\end{algorithmic}
\end{algorithm}

\section{Experiments}\label{sec:experiments}

\subsection{Experimental Settings}

\begin{table}[t!]
    \caption{Evaluation of models across all datasets for disentanglement (DCI and InfoMEC) and reconstruction (MSE $\times 10^4$) performance.}
    \label{tab:evaluation_table}
    \centering
    \scriptsize
    \begin{tabular}{lccccccc}
        \toprule
        model & \multicolumn{1}{c}{$\mathsf{D} \uparrow$} & \multicolumn{1}{c}{$\mathsf{I} \uparrow$} & \multicolumn{1}{c}{$\mathsf{C} \uparrow$} & \multicolumn{1}{c}{InfoM $\uparrow$} & \multicolumn{1}{c}{InfoE $\uparrow$} & \multicolumn{1}{c}{InfoC $\uparrow$} & \multicolumn{1}{c}{MSE $\downarrow$} \\
        \midrule
        &\multicolumn{7}{c}{Shapes3D} \\
        \cmidrule{2-8}
        AE & 0.52 & {\color{gray} 0.82} & {\color{lightgray} 0.42} & 0.55 & {\color{gray} \textbf{0.99}} & {\color{lightgray} 0.41} & \textbf{0.3} \\

$\beta$-VAE & 0.56 & {\color{gray} 0.89} & {\color{lightgray} 0.65} & 0.62 & {\color{gray} 0.76} & {\color{lightgray} 0.75} & 1.6 \\

FactorVAE & 0.72 & {\color{gray} 0.94} & {\color{lightgray} \textbf{0.79}} & 0.61 & {\color{gray} 0.81} & {\color{lightgray} \textbf{0.86}} & 1.6 \\

BioAE & 0.41 & {\color{gray} 0.75} & {\color{lightgray} 0.32} & 0.52 & {\color{gray} \textbf{0.99}} & {\color{lightgray} 0.33} & 1.1 \\

$\alpha$-TCVAE & 0.60 & {\color{gray} 0.79} & {\color{lightgray} 0.37} & 0.51 & {\color{gray} 0.72} & {\color{lightgray} 0.36} & 1.3\\

QLAE & \textbf{0.91} & {\color{gray} \textbf{1.00}} & {\color{lightgray} 0.75} & 0.71 & {\color{gray} \textbf{0.99}} & {\color{lightgray} 0.54} & 0.9 \\

dVAE & 0.89 & {\color{gray} 0.95} & {\color{lightgray} 0.71} & 0.70 & {\color{gray} \textbf{0.99}} & {\color{lightgray} 0.45} & 4.4 \\

QVAE & 0.73 & {\color{gray} 0.94} & {\color{lightgray} 0.54} & 0.69 & {\color{gray} 0.93} & {\color{lightgray} 0.38} & 0.8 \\

FactordVAE (ours) & {\color{darkgray} 0.91} & {\color{gray} 0.95} & {\color{lightgray} 0.70} & 0.49 & {\color{gray} \textbf{0.99}} & {\color{lightgray} 0.32} & 4.7 \\

FactorQVAE (ours) & 0.86 & {\color{gray} 0.97} & {\color{lightgray} 0.65} & \textbf{0.84} & {\color{gray} 0.89} & {\color{lightgray} 0.52} & {\color{darkgray} \textbf{0.6}}
 \\
        \midrule
        &\multicolumn{7}{c}{Isaac3D} \\
        \cmidrule{2-8}
        AE & 0.22 & {\color{gray} 0.75} & {\color{lightgray} 0.18} & 0.39 & {\color{gray} 0.61} & {\color{lightgray} 0.15} & \textbf{0.2} \\

$\beta$-VAE & 0.45 & {\color{gray} 0.80} & {\color{lightgray} 0.58} & 0.49 & {\color{gray} 0.46} & {\color{lightgray} 0.53} & 1.5 \\

FactorVAE & 0.51 & {\color{gray} 0.80} & {\color{lightgray} \textbf{0.64}} & 0.56 & {\color{gray} 0.47} & {\color{lightgray} \textbf{0.62}} & 1.6 \\

BioAE & 0.29 & {\color{gray} 0.78} & {\color{lightgray} 0.23} & 0.40 & {\color{gray} 0.57} & {\color{lightgray} 0.17} & 0.8 \\

$\alpha$-TCVAE & 0.37 & {\color{gray} 0.73} & {\color{lightgray} 0.26} & 0.45 & {\color{gray} 0.44} & {\color{lightgray} 0.19} & 1.8 \\

QLAE & 0.57 & {\color{gray} \textbf{0.89}} & {\color{lightgray} 0.48} & 0.52 & {\color{gray} 0.65} & {\color{lightgray} 0.37} & 0.5 \\

dVAE & 0.53 & {\color{gray} 0.80} & {\color{lightgray} 0.45} & 0.60 & {\color{gray} 0.62} & {\color{lightgray} 0.33} & 3.6 \\

QVAE & 0.56 & {\color{gray} 0.81} & {\color{lightgray} 0.49} & 0.51 & {\color{gray} \textbf{0.72}} & {\color{lightgray} 0.34} & {\color{darkgray} \textbf{0.4}} \\

FactordVAE (ours) & {\color{darkgray} 0.62} & {\color{gray} 0.78} & {\color{lightgray} 0.56} & {\color{darkgray} 0.86} & {\color{gray} 0.66} & {\color{lightgray} 0.56} & 5.7 \\

FactorQVAE (ours) & \textbf{0.59} & {\color{gray} 0.75} & {\color{lightgray} 0.56} & \textbf{0.68} & {\color{gray} 0.64} & {\color{lightgray} 0.45} & 0.7
 \\
        \midrule
        &\multicolumn{7}{c}{MPI3D} \\
        \cmidrule{2-8}
        AE & 0.10 & {\color{gray} 0.57} & {\color{lightgray} 0.10} & 0.30 & {\color{gray} 0.25} & {\color{lightgray} 0.11} & \textbf{0.5} \\

$\beta$-VAE & 0.36 & {\color{gray} 0.70} & {\color{lightgray} 0.42} & 0.39 & {\color{gray} 0.35} & {\color{lightgray} \textbf{0.39}} & 1.2 \\

FactorVAE & 0.20 & {\color{gray} 0.61} & {\color{lightgray} 0.31} & \textbf{0.40} & {\color{gray} 0.23} & {\color{lightgray} 0.32} & 1.4 \\

BioAE & 0.23 & {\color{gray} 0.66} & {\color{lightgray} 0.26} & 0.28 & {\color{gray} 0.37} & {\color{lightgray} 0.24} & 0.9 \\

$\alpha$-TCVAE & 0.20 & {\color{gray} 0.78} & {\color{lightgray} 0.28} & 0.31 & {\color{gray} 0.29} & {\color{lightgray} 0.26} & 1.5 \\

QLAE & 0.36 & {\color{gray} \textbf{0.76}} & {\color{lightgray} 0.41} & 0.31 & {\color{gray} 0.40} & {\color{lightgray} 0.38} & {\color{darkgray} \textbf{0.6}} \\

dVAE & 0.33 & {\color{gray} 0.67} & {\color{lightgray} 0.31} & 0.39 & {\color{gray} 0.66} & {\color{lightgray} 0.23} & 2.2 \\

QVAE & 0.34 & {\color{gray} 0.61} & {\color{lightgray} 0.31} & 0.31 & {\color{gray} 0.77} & {\color{lightgray} 0.27} & 0.9 \\

FactordVAE (ours) & 0.29 & {\color{gray} 0.56} & {\color{lightgray} 0.23} & 0.37 & {\color{gray} 0.60} & {\color{lightgray} 0.25} & 4.4 \\

FactorQVAE (ours) & {\textbf{0.46}} & {\color{gray} 0.67} & {\color{lightgray} \textbf{0.44}} & 0.33 & {\color{gray} \textbf{0.94}} & {\color{lightgray} 0.36} & 1.1
 \\
        \bottomrule
    \end{tabular}
\end{table}

We conduct our experiments on three datasets: Shapes3D \cite{burgess20183d}, Isaac3D \cite{nie2019high}, and MPI3D \cite{gondal2019transfer}. Each dataset corresponds to simple, medium, hard complexity, respectively. In our experiments, MPI3D refers to the MPI3D-Complex variant, which contains images captured from a real robotic arm and physical objects under varying backgrounds and lighting conditions, making it representative of real-world scenarios. Further details about the datasets are given in Appendix~\ref{sec:datasets}.

We choose fundamental, recent, and state-of-the-art disentanglement models to compare our model: $\beta$-VAE \cite{higgins2017beta}, FactorVAE \cite{kim2018disentangling}, BioAE \cite{whittington2023disentanglement}, $\alpha$-TCVAE \cite{meo2024alpha}, and QLAE \cite{hsu2023disentanglement}. We also evaluate autoencoder (AE), QVAE which is the scalar quantizing version of VQVAE \cite{oord2017neural} instead of vector quantization, and dVAE \cite{ramesh2021zeroshot}. We use the same network architecture across all baseline models to ensure a fair comparison. While this may result in minor deviations from the original findings reported in the literature, these differences are not significant enough to impact the overall conclusions. Further details, such as hyperparameter selection and ablations, are provided in Appendix~\ref{sec:hyperparameters}, and the network architecture is described in Appendix~\ref{sec:model_description}.

\subsection{Discussions}

\begin{figure}[t!]
    \centering
    \begin{subfigure}[b]{0.45\textwidth}
        \centering
        \includegraphics[width=\textwidth]{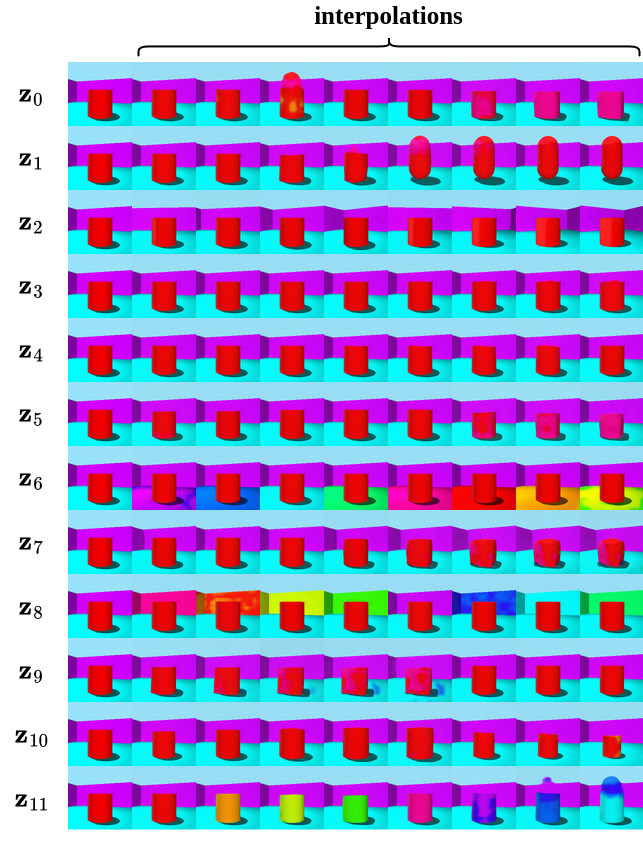}     
        \caption{QLAE}
    \end{subfigure}
    \hfill
    \begin{subfigure}[b]{0.45\textwidth}
        \centering
        \includegraphics[width=\textwidth]{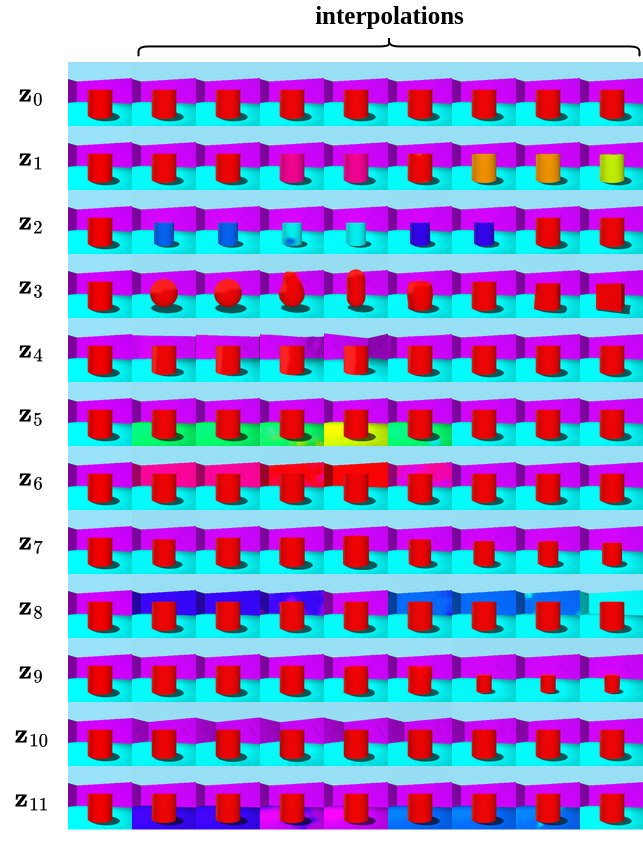}
        \caption{FactorQVAE}
    \end{subfigure}
    \caption{Latent traversal on the same image with QLAE and FactorQVAE for Shapes3D dataset. Each row $i$ (labeled as $\mathbf{z_i}$) shows the result of manipulating the $i^{th}$ latent, with the last 8 columns showing interpolations. For QLAE, the $i^{th}$ latent is intervened on with a linear interpolation between the minimum and maximum values in the corresponding $i^{th}$ codebook while it is intervened on with a linear interpolation between the minimum and maximum values in the global codebook for FactorQVAE. For FactorQVAE, rows 1 and 2 control object hue, row 3 controls object shape, rows 4 and 10 control camera orientation, rows 5 and 11 control floor hue, rows 6 and 8 control wall hue, and rows 7 and 9 control object scale.}
    \label{fig:shapes3d_traversal}
\end{figure}

\begin{figure}[t!]
    \centering
    \begin{subfigure}[b]{0.45\textwidth}
        \centering
        \includegraphics[width=\textwidth]{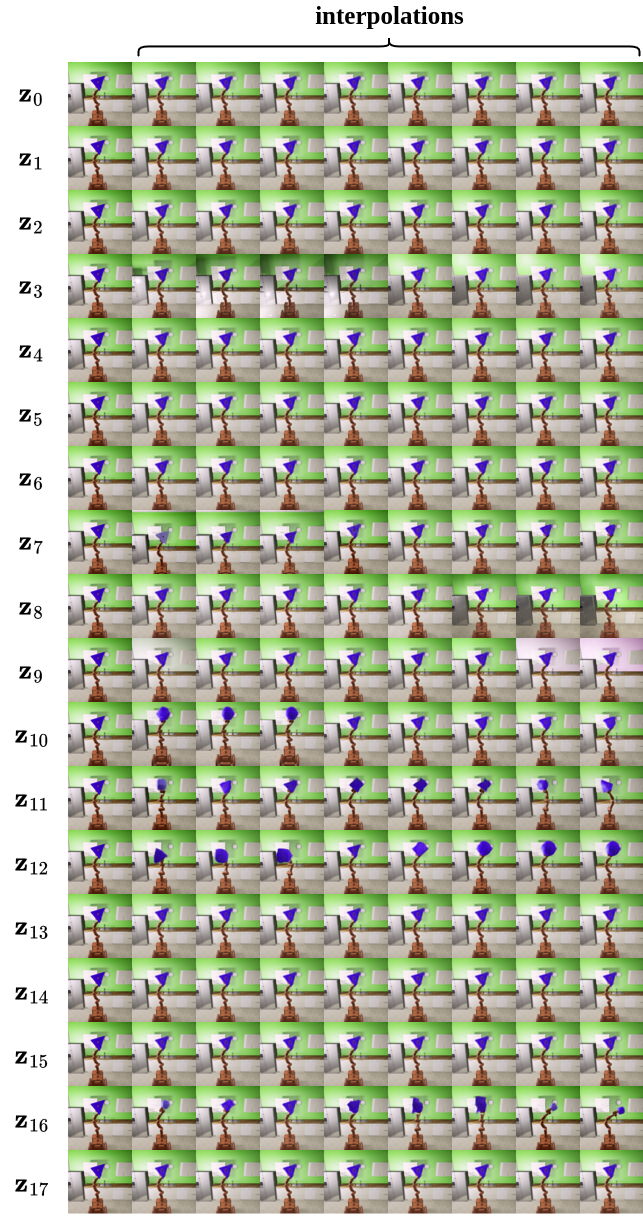}     
        \caption{FactorVAE}
    \end{subfigure}
    \hfill
    \begin{subfigure}[b]{0.45\textwidth}
        \centering
        \includegraphics[width=\textwidth]{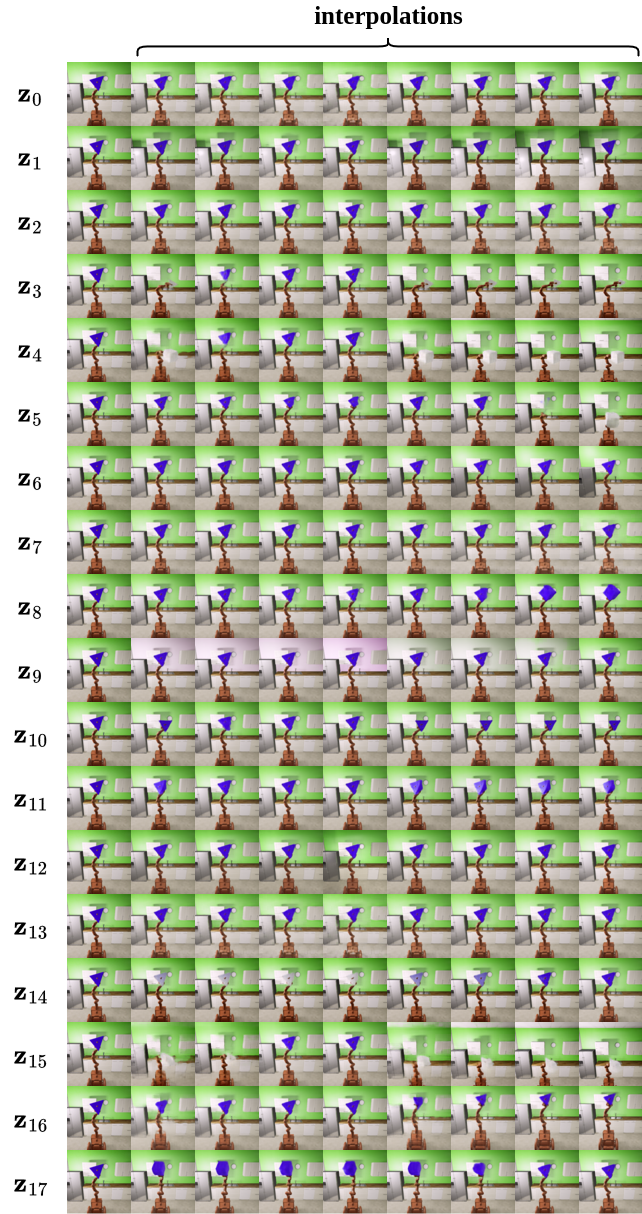}
        \caption{FactorQVAE}
    \end{subfigure}
    \caption{Latent traversal on the same image with FactorVAE and FactorQVAE for Isaac3D dataset. Each row $i$ (labeled as $\mathbf{z_i}$) shows the result of manipulating the $i^{th}$ latent, with the last 8 columns showing interpolations. For FactorVAE, the $i^{th}$ latent is intervened on with a linear interpolation between "original latent value - 3" and "original latent value + 3" while it is intervened on with a linear interpolation between the minimum and maximum values in the global codebook for FactorQVAE. For FactorQVAE, rows 1 and 6 control lighting direction, rows 3 and 10 control the robot’s vertical axis, rows 5 and 8 control object scale, rows 9 and 13 control wall color, rows 10 and 14 control object color, row 11 controls the robot’s horizontal axis, row 12 controls lighting intensity, row 16 controls camera height, and row 17 controls object shape.}
    \label{fig:isaac3d_traversal}
\end{figure}

\begin{figure}[t!]
    \centering
    \begin{subfigure}[b]{0.45\textwidth}
        \centering
        \includegraphics[width=\textwidth]{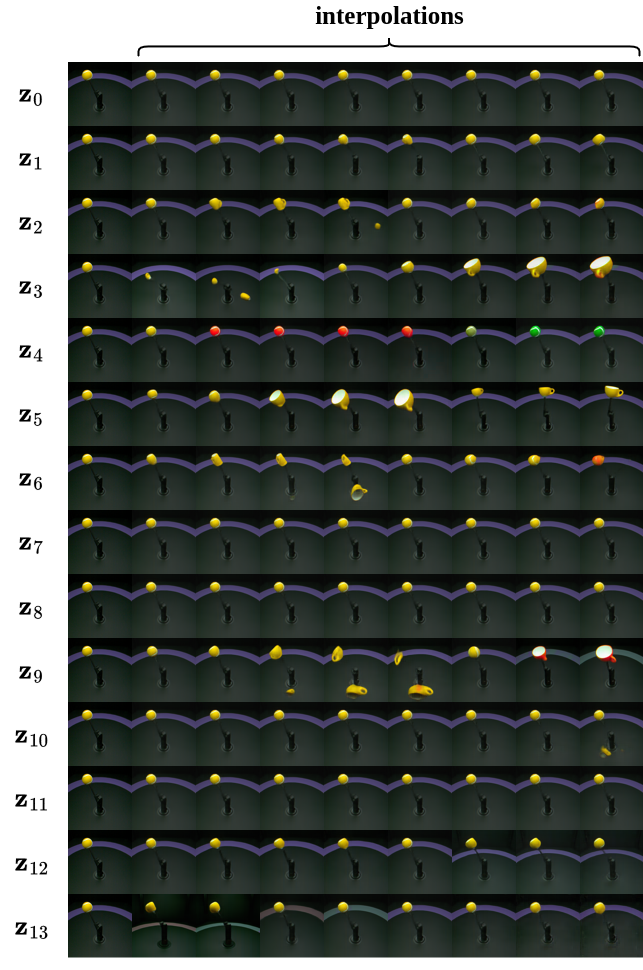}     
        \caption{$\beta$-VAE}
    \end{subfigure}
    \hfill
    \begin{subfigure}[b]{0.45\textwidth}
        \centering
        \includegraphics[width=\textwidth]{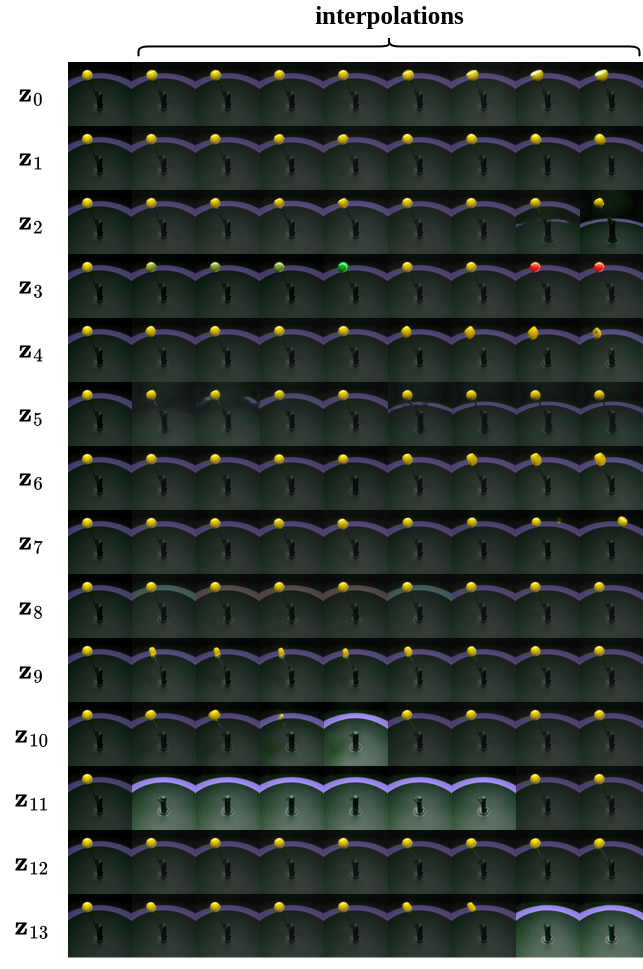}
        \caption{FactorQVAE}
    \end{subfigure}
    \caption{Latent traversal on the same image with $\beta$-VAE and FactorQVAE for MPI3D dataset. Each row $i$ (labeled as $\mathbf{z_i}$) shows the result of manipulating the $i^{th}$ latent, with the last 8 columns showing interpolations. For $\beta$-VAE, the $i^{th}$ latent is intervened on with a linear interpolation between "original latent value - 3" and "original latent value + 3" while it is intervened on with a linear interpolation between the minimum and maximum values in the global codebook for FactorQVAE. For FactorQVAE, row 2 controls camera height, row 3 controls object color, row 5 controls object size, row 8 controls background color, and row 9 controls object shape.}
    \label{fig:mpi3d_traversal}
\end{figure}

\begin{figure}[t!]
    \centering
    \includesvg[width=\textwidth]{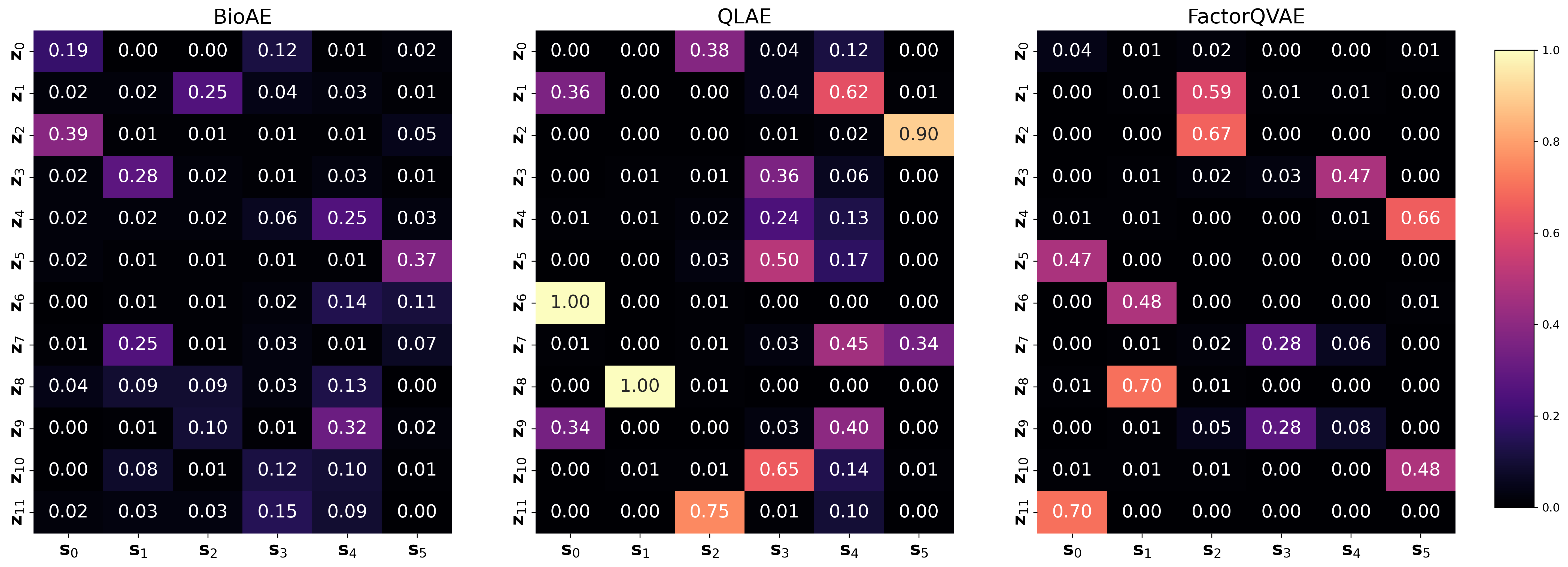}
    \caption{Visualization of the NMI matrix which is used in InfoMEC calculation for Shapes3D dataset.}
    \label{fig:shapes3d_nmi}
\end{figure}

\begin{figure}[t!]
    \centering
    \includesvg[width=\textwidth]{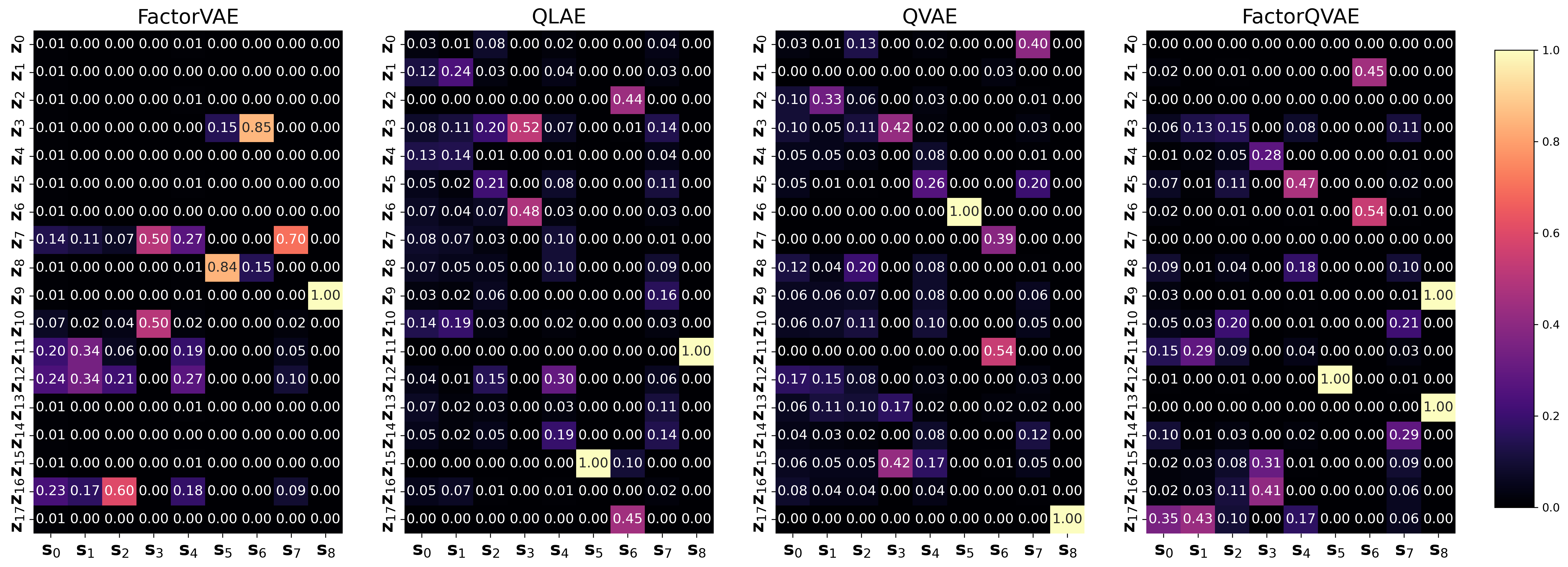}
    \caption{Visualization of the information matrices used in DCI calculation for Isaac3D dataset.}
    \label{fig:isaac3d_importance}
\end{figure}

\begin{figure}[t!]
    \centering
    \includesvg[width=\textwidth]{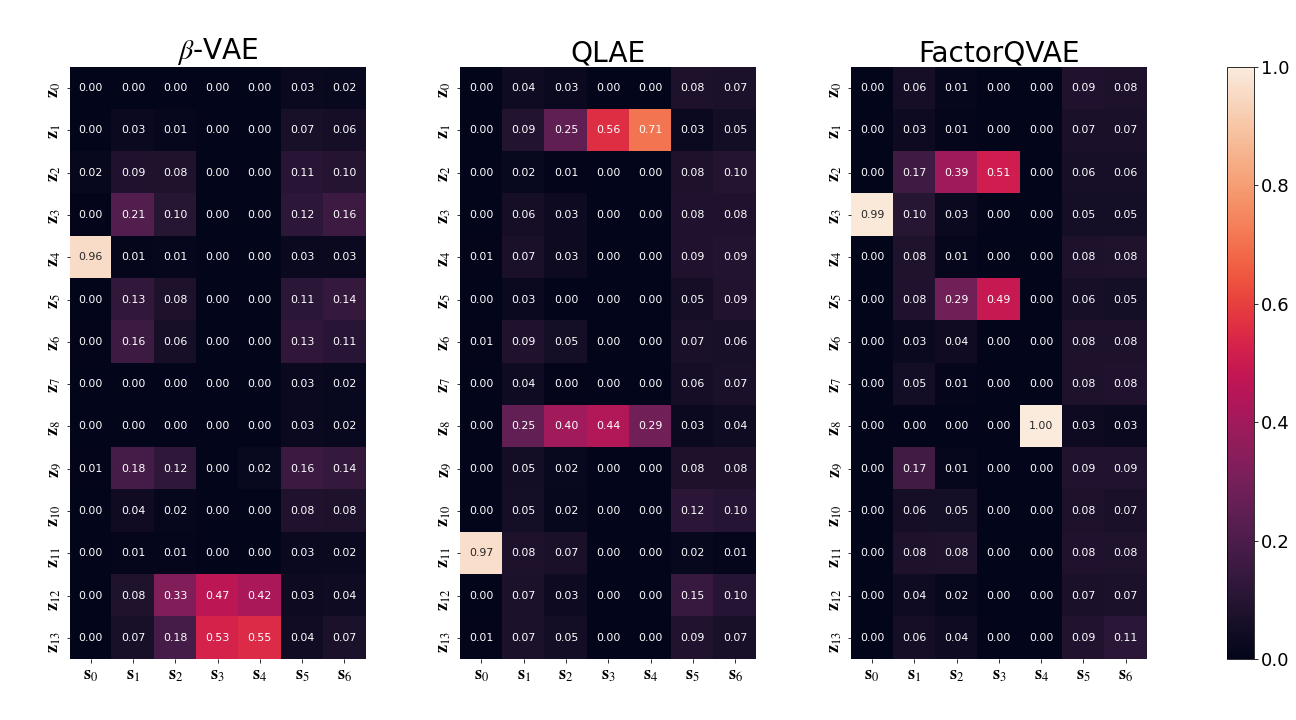}
    \caption{Visualization of the information matrix which is used in DCI calculation for MPI3D dataset.}
    \label{fig:mpi3d_importance}
\end{figure}

\begin{figure}[t!]
    \centering
    \includegraphics[width=\textwidth]{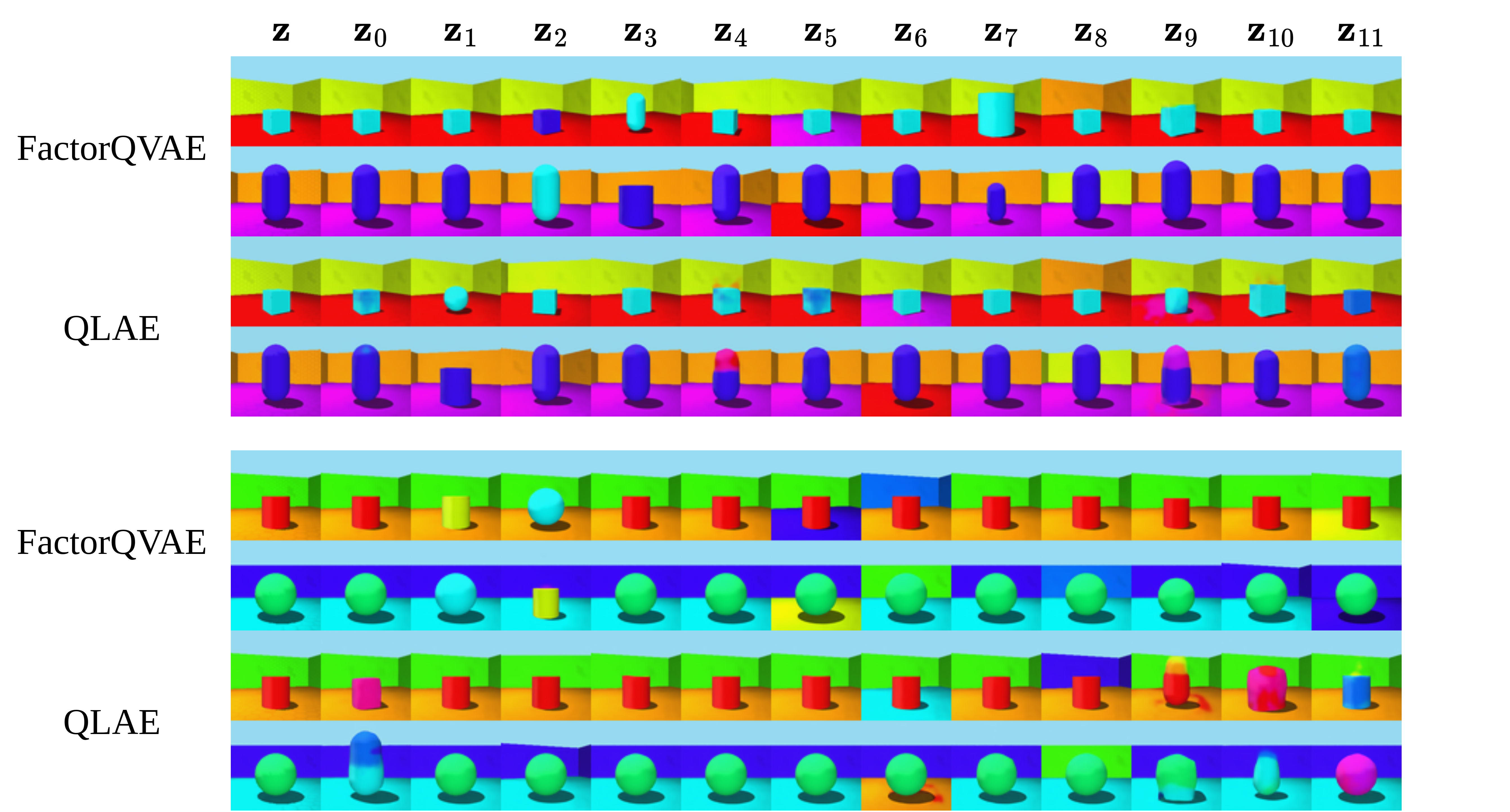}
    \caption{Latent variable swapping between two different images. We select two different pairs of images, and show them in column $\mathbf{z}$ of each image block. For each image block, the first two rows show the latent swapping results for FactorQVAE while the last two rows show the latent swapping results for QLAE. Each $\mathbf{z_i}$ column displays the effects of swapping the $i^{th}$ latent variables between two images.}
    \label{fig:shapes3d_swap}
\end{figure}

We conduct various experiments as ablation studies to analyze the effects of discrete representation learning and factorization separately. We detail the results of selecting discrete representation learning over continuous representation learning, QVAE over dVAE, scalar quantization over vector quantization, and a global codebook over codebooks per latent dimension in this section. We further comment on the contribution of each design choice to the overall performance of our model FactorQVAE based on our ablation studies.

To evaluate the disentanglement numerically, we use DCI and InfoMEC, with details provided in Section~\ref{sec:background_metrics}. Table~\ref{tab:evaluation_table} shows both DCI and InfoMEC results of the models for all datasets. As disentanglement should be obtained along with acceptable reconstruction performance, we further present the mean squared error (MSE) of the models in Table~\ref{tab:evaluation_table}. To highlight our key results, we draw attention to the $\mathsf{D}$ and InfoM scores in Table~\ref{tab:evaluation_table}, where our method clearly outperforms the baselines in most of the cases. While $\mathsf{C}$ and InfoC scores may not show a similar level of improvement, this aligns with our discussion in Section~\ref{sec:background_metrics}, where we prioritize disentanglement (modularity) to completeness (compactness) due to the inherent trade-off dictated by the chosen parameter settings ($d = 2*F$).

For the visual evaluation of disentanglement and reconstruction performance, we intervene on the latent variables separately, and decode it to the image space to see if only a single generative factor is affected without artifacts in the image.

Figure~\ref{fig:shapes3d_traversal} compares FactorQVAE with its closest alternative, QLAE, on Shapes3D. While QLAE still struggles with image reconstruction artifacts, FactorQVAE produces cleaner results. Additionally, QLAE’s outputs are more entangled, as seen in the 8th row, where \textit{orientation} and \textit{shape} are intertwined. Although FactorQVAE shows fewer artifacts and greater disentanglement, some values for generative factors are still missing. For instance, in the 6th and 12th rows, \textit{orange} is absent from the \textit{floor color}, suggesting that a generative factor’s value still depends on the semantic information from other factors. This highlights the ongoing challenge of achieving full disentanglement, even with a simpler dataset. Figure~\ref{fig:isaac3d_traversal} compares the results of FactorQVAE and FactorVAE on the Isaac3D dataset. FactorQVAE produces cleaner reconstructions with fewer artifacts and shows more consistent disentanglement. In FactorVAE, we observe that changing some latent variables affects multiple generative factors, whereas FactorQVAE exhibits more stable and independent control of the latent variables. Figure~\ref{fig:mpi3d_traversal} present the image reconstruction results after latent traversals for MPI3D datasets. While we observe that FactorQVAE and the other models yield better disentanglement for Shapes3D and Isaac3D datasets in Figure~\ref{fig:shapes3d_traversal} and Figure~\ref{fig:isaac3d_traversal}, respectively, Figure~\ref{fig:mpi3d_traversal} clearly demonstrates the challenges of disentangling MPI3D dataset for both $\beta$-VAE and FactorQVAE. Therefore, we observe that there is still a long way to achieve an acceptable level of disentanglement in challenging environments.

Figure~\ref{fig:shapes3d_nmi} presents the heatmaps of NMI matrices used in InfoMEC calculation for Shapes3D, and Figure~\ref{fig:isaac3d_importance} and Figure~\ref{fig:mpi3d_importance} present the heatmaps of the information matrices used in DCI calculation for Isaac3D and MPI3D, respectively. As we use different disentanglement metrics DCI and InfoMEC, we visually showcase how they are calculated. Figure~\ref{fig:shapes3d_nmi} and Figure~\ref{fig:isaac3d_importance} clearly demonstrate the relationship between the latent variables and the generative sources, with FactorQVAE appearing to balance disentanglement and completeness better than the other models. On the other hand, Figure~\ref{fig:mpi3d_importance} visually demonstrates the failure of all models in capturing all the generative factors in a challenging dataset like MPI3D. Still, FactorQVAE seems to be capturing more factors in a disentangled manner.

We conduct an additional experiment on the Shapes3D dataset, where we swap the latent variables of two images individually. Figure~\ref{fig:shapes3d_swap} presents the results of this latent variable swapping for QLAE and FactorQVAE, using two pairs of images. \textbf{z} column shows the original images, and the $\mathbf{z_i}$ column displays the results after swapping the $i^{th}$ latent variables. QLAE shows poor reconstruction performance, with noticeable artifacts, particularly for latent variables $\mathbf{z_0}$ and $\mathbf{z_9}$, which do not correspond to a single generative factor (see Figure~\ref{fig:shapes3d_nmi} for the relationship between latent variables and generative sources). In contrast, FactorQVAE successfully transfers the generative factor values between images without introducing artifacts.

For the first image pair, the generative factors are swapped exactly between the images. However, when multiple latent variables represent the same generative factor, such as $\mathbf{z_5}$ and $\mathbf{z_{11}}$ for \textit{floor hue}, only one may influence the reconstructions. For the second image pair, the exact values of the generative factors may not transfer perfectly, as seen in the \textit{floor hue} swapping with $\mathbf{z_5}$ and $\mathbf{z_{11}}$. This result highlights that, even if a latent variable represents a single generative factor, its effect may still depend on the image’s semantic content.

\textbf{Discrete vs continuous representation learning:} To start with, we compare discrete representation learning and continuous representation learning in terms of disentanglement and reconstruction. By comparing discrete representation models (FactorQVAE, QLAE, and QVAE) against continuous representation methods ($\beta$-VAE, FactorVAE, and $\alpha$-TCVAE) in Table~\ref{tab:evaluation_table}, we observe that discrete latent models consistently outperform continuous methods in both DCI and InfoMEC metrics across datasets. In addition to disentanglement performance, the discrete representation learning models achieve better reconstruction performance. Since achieving both acceptable reconstruction and disentanglement performance together is difficult, an AE trained solely on the reconstruction objective provides the best reconstruction performance, as expected. However, discrete representation learning models can also achieve reconstruction performance that is acceptably close to that of autoencoders (AE).

\textbf{QVAE vs dVAE:} Although discrete representation learning improves performance, the design of the model remains crucial for this enhancement. Therefore, we implement dVAE and QVAE as the discrete representation learning models, and observe that QVAE leads to more balanced performance between reconstruction and disentanglement. When we go through the disentanglement performance of dVAE and QVAE in Table~\ref{tab:evaluation_table}, we see that dVAE outperforms QVAE in most of the settings. However, dVAE cannot achieve adequate reconstruction performance as it seems, which is essential along with disentanglement. 

We can view the issue with dVAE as a posterior collapse problem, meaning that the model prioritizes minimizing the KL term rather than the reconstruction term. As a result, the latent representation fails to sufficiently capture information about the input data, although it achieves better disentanglement. This behavior likely stems from the fact that the parameters of the Categorical distribution are learned by the encoder, which is more challenging to optimize. Consequently, the model finds a shortcut to minimize the loss by focusing on reducing the KL term instead of maintaining a proper balance. Given this performance gap between QVAE and dVAE, we proceed with QVAE.

\textbf{Effects of factorization:} After demonstrating the contribution of discrete representation learning in terms of both disentanglement and reconstruction performance, we incorporate factorization as an inductive bias to further enhance disentanglement. Our model FactorQVAE achieves the best performance in most of the settings for both DCI and InfoMEC as observed in Table~\ref{tab:evaluation_table}. As stated in Section~\ref{sec:background_metrics}, the disentanglement ($\mathsf{D}$) and its counterpart, modularity (InfoM), are prioritized to evaluate the disentanglement in general. We observe that FactorQVAE achieves the best $\mathsf{D}$ and InfoM values in most of the settings while it also achieves compatible completeness (compactness) values. Besides, FactorQVAE's reconstruction performance is on a par with the models having the best reconstruction performance. 

We further highlight the FactorVAE results to emphasize on the effects of factorization in continuous representation learning. We present that FactorVAE generally outperforms $\beta$-VAE in terms of disentanglement while achieving similar reconstruction performance with the latter. Moreover, FactorVAE achieves the best DCI and InfoMEC scores in some of the settings, demonstrating the effectiveness of factorization. Therefore, combining factorization with discrete representation learning boosts the performance as expected. 

Lastly, we observe that factorization also improves the performance of dVAE in terms of disentanglement. Even though FactordVAE achieves the best DCI and InfoMEC scores in some of the settings, FactordVAE is not favorable as a consequence of its poor reconstruction performance. 

\begin{table}[t!]
    \caption{Effects of codebook ($\mathcal{M}$) design using Shapes3D dataset.}
    \label{tab:codebook_design}
    \centering
    \scriptsize
    \begin{tabular}{lcccccccc}
        \toprule
        $\mathcal{M}$ & \multicolumn{1}{c}{$\mathsf{D} \uparrow$} & \multicolumn{1}{c}{$\mathsf{I} \uparrow$} & \multicolumn{1}{c}{$\mathsf{C} \uparrow$} & \multicolumn{1}{c}{InfoM $\uparrow$} & \multicolumn{1}{c}{InfoE $\uparrow$} & \multicolumn{1}{c}{InfoC $\uparrow$} & \multicolumn{1}{c}{MSE $\downarrow$} \\
        \midrule
        & \multicolumn{7}{c}{(V)QVAE} \\
        \cmidrule{2-8}
        $32 \times 1$ & 0.69 & {\color{gray} 0.97} & {\color{lightgray} 0.52} & 0.75 & {\color{gray} 0.88} & {\color{lightgray} 0.39} & 1.4 \\

$64 \times 1$ & 0.73 & {\color{gray} 0.94} & {\color{lightgray} 0.54} & 0.69 & {\color{gray} 0.93} & {\color{lightgray} 0.38} & 0.8 \\

$128 \times 1$ & 0.72 & {\color{gray} 0.79} & {\color{lightgray} 0.64} & 0.51 & {\color{gray} 0.98} & {\color{lightgray} 0.41} & 1.7 \\

$64 \times 8$ & 0.68 & {\color{gray} 0.99} & {\color{lightgray} 0.50} & 0.52 & {\color{gray} 0.99} & {\color{lightgray} 0.33} & 0.6 \\

$64 \times 16$ & 0.65 & {\color{gray} 0.99} & {\color{lightgray} 0.49} & 0.47 & {\color{gray} 0.99} & {\color{lightgray} 0.29} & 0.4 \\
        \midrule
        & \multicolumn{7}{c}{Factor(V)QVAE} \\
        \cmidrule{2-8}
        $32 \times 1$ & 0.81 & {\color{gray} 0.94} & {\color{lightgray} 0.66} & 0.82 & {\color{gray} 0.92} & {\color{lightgray} 0.57} & 1.1 \\

$64 \times 1$ & 0.86 & {\color{gray} 0.97} & {\color{lightgray} 0.65} & 0.84 & {\color{gray} 0.89} & {\color{lightgray} 0.52} & 0.6 \\

$128 \times 1$ & 0.82 & {\color{gray} 0.91} & {\color{lightgray} 0.62} & 0.68 & {\color{gray} 0.97} & {\color{lightgray} 0.41} & 1.5 \\

$64 \times 8$ & 0.78 & {\color{gray} 0.97} & {\color{lightgray} 0.63} & 0.59 & {\color{gray} 0.99} & {\color{lightgray} 0.48} & 0.5 \\

$64 \times 16$ & 0.75 & {\color{gray} 0.96} & {\color{lightgray} 0.60} & 0.51 & {\color{gray} 0.98} & {\color{lightgray} 0.45} & 0.4 \\
        \bottomrule
  \end{tabular}
\end{table}

\textbf{Scalar quantization vs vector quantization:} To substantiate the methodological choice detailed in Section~\ref{sec:method}, we conduct an ablation study comparing scalar and vector quantization strategies using the Shapes3D dataset. Results in Table~\ref{tab:codebook_design} demonstrate that vector quantization marginally improves reconstruction but substantially reduces disentanglement scores (e.g., InfoMEC metric). Additionally, we observe increased training instability in vector quantized models (e.g., VQVAE and FactorVQVAE with $\mathcal{M} \in \mathbb{R}^{64 \times 8}$), which are more sensitive to hyperparameter variations. These findings confirm our hypothesis that scalar quantization provides a more suitable trade-off, effectively balancing disentanglement and reconstruction performance.

Beyond scalar and vector quantization comparison, we look at the effects of number of elements in $\mathcal{M}$ for scalar quantization. We see a slight difference in evaluation metrics when we have different number of elements in $\mathcal{M}$. That indicates that the performance of both QVAE and FactorQVAE is resistant to number of elements in $\mathcal{M}$. Therefore, we prefer not to engineer dataset specific values for the number of elements in $\mathcal{M}$, and proceed with 64 number of scalar values in $\mathcal{M}$ for all datasets.

\begin{figure}[t!]
    \centering
    \includegraphics[width=\textwidth]{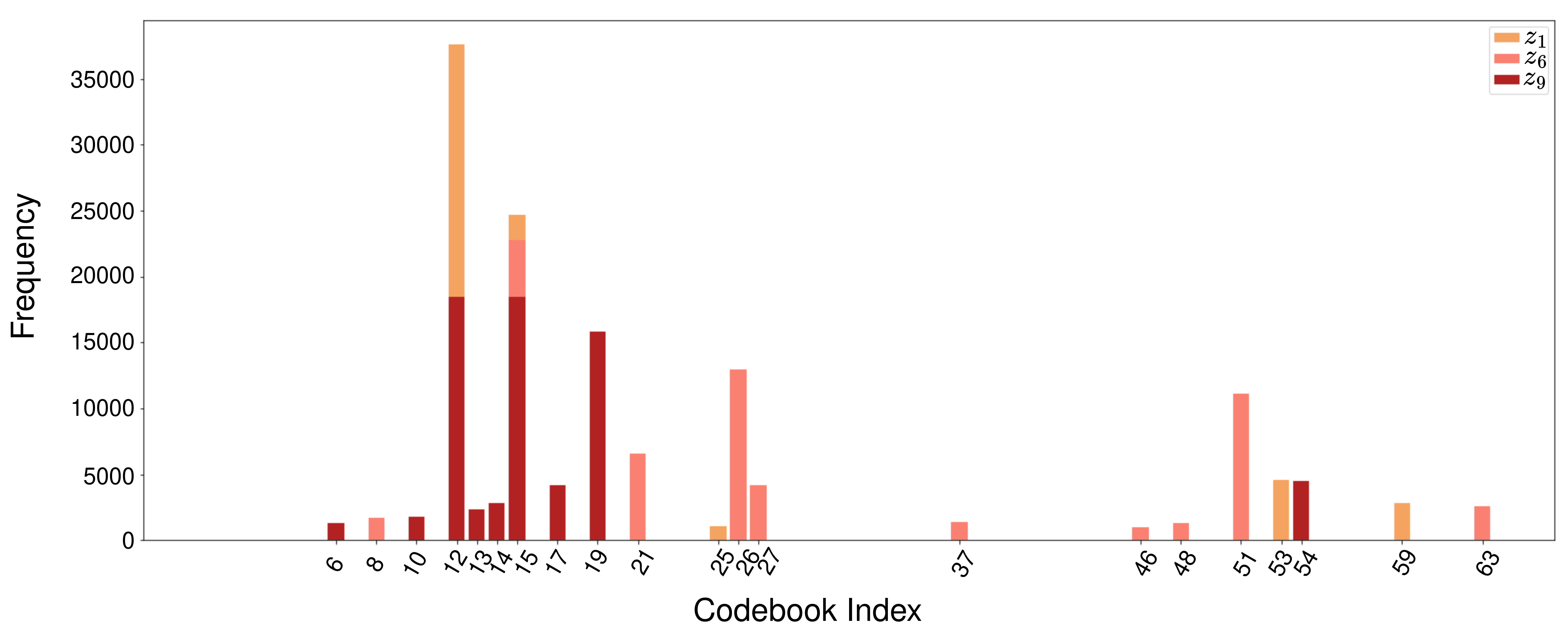}
    \caption{Frequencies of the codebook elements used in specific latent dimensions learned by FactorQVAE in Isaac3D dataset.}
    \label{fig:histogram}
\end{figure}

\textbf{Global codebook vs codebooks per latent dimensions:} The efficiency of using a global codebook instead of codebooks per latent dimensions can be observed to some extent by comparing QVAE with QLAE. Even though QLAE uses a deterministic categorical posterior for each latent variable rather than a stochastic categorical posterior like ours, we can still make sense out of this comparison. 

When we look at Table~\ref{tab:evaluation_table}, we see that QVAE is on a par with QLAE both for disentanglement and reconstruction performance, which does not highlight the significance of having a global codebook. However, the good side of having a global codebook is that we let the model learn how to partition the latent space. As we exemplify in Section~\ref{sec:introduction}, different generative factors might have different number of options. Hence, setting fixed length codebooks per latent might be restrictive in terms of representativeness. To support our intuition, we conduct an experiment on Isaac3D dataset for FactorQVAE. Figure~\ref{fig:isaac3d_importance} shows that for FactorQVAE, latents $z_1$ and $z_6$ capture the generative factor $s_6$, while $z_9$ captures $s_8$. As detailed in Section~\ref{sec:datasets}, $s_6$ is the lighting direction, and it has 6 possible values, while $s_8$ is wall color, and it has 4 possible values. Based on our intuition, $s_6$ should be represented with a latent space larger than $s_8$'s latent space. Thus, we analyze which codebook elements are used in the latent dimensions capturing these generative sources, and their frequencies, in Figure~\ref{fig:histogram}. We observe that the number of codebook elements representing $s_6$ in latent dimensions $z_1$ and $z_6$ is greater than the number of codebook elements representing $s_8$ in $z_9$, which aligns with our expectations. We further observe in Figure~\ref{fig:histogram} that the sets of codebook elements representing different generative factors are mostly disjoint, serving as an evidence that FactorQVAE learns to assign different sets of codebook elements for different generative factors. The overlapping codebook elements for different generative factors capture more information, and are used in various latent dimensions. Even though we still cannot talk about a perfect separation of the codebook, our model provides a promising direction on this matter.

\textbf{Comments on the contribution of each component:} Considering our detailed comparisons about every design choice and the evaluation results in Table~\ref{tab:evaluation_table}, we believe that looking at the results of $\beta$-VAE vs. FactorVAE and QLAE/QVAE vs. FactorQVAE highlight the contribution of factorization while $\beta$-VAE vs. QLAE/QVAE and FactorVAE vs. FactorQVAE comparisons demonstrate the contribution of discretization to the performance. We draw a conclusion from these ablation studies that the discrete representation learning has the biggest positive effect on the performance while factorization seems likewise essential for the disentanglement with both continuous and discrete representation learning. Therefore, we experimentally conclude that FactorQVAE consists of all essential design properties that neither of them are replaceable for better disentanglement.

\section{Conclusion}\label{sec:conclusion}

The proposed FactorQVAE extends the discrete representation learning model QVAE by introducing a regularizer called factorization to enhance disentanglement in representation learning. First, we show that discrete representation learning is more suitable than continuous representation learning for disentanglement. We further demonstrate that incorporating an inductive bias into a discrete representation learning model improves its performance, with FactorQVAE outperforming previous methods in most settings in terms of both disentanglement and reconstruction. Since our evaluation focuses on standard 3D-object benchmarks in disentanglement research, the conclusions are primarily applicable to structured visual domains with well-defined generative factors. 3D scenes are a particularly suitable testbed since their generative factors are naturally interpretable and controllable, making them central to disentanglement research. Extending FactorQVAE to less structured datasets or non-visual modalities remains a promising direction for future work.

Although the proposed method improves disentanglement performance, there is still room for improvement in challenging settings, such as real-world environments exemplified by the MPI3D dataset. We observe that neither previous methods nor FactorQVAE achieve an acceptable level of disentangled representation learning in such challenging settings. We conjecture that this limitation can be attributed to the fact that real-world environments consist of intricate and inter-dependent generative factors which cannot be handled by just assuming \textit{i)} fully continuous or fully discrete representations and \textit{ii)} full independence between the generative factors. Therefore, we believe our work can inspire future developments that combine discrete representations with continuous representations to address disentangled representation learning for real-world problems, as we have already demonstrated the effectiveness of discrete representation learning for disentanglement within a factorizable scalar latent space.

\section*{Acknowledgment}
In this work, Gulcin Baykal was supported by TÜB\.{I}TAK 2214-A International Research Fellowship Programme for PhD Students and Google DeepMind Scholarship Program at ITU. Computing resources used during this research were provided by the Scientific Research Project Unit of Istanbul Technical University [project number MOA-2019-42321].

\appendix
\section*{Appendix}
\renewcommand{\thesection}{\Alph{section}}

\section{Further Experimental Details}

We use PyTorch Lightning \citep{falcon2019lightning} framework in our implementation. All models are trained for 100K iterations on every dataset using a single NVIDIA V100 GPU. We use a batch size of 256 and the Adam optimizer with an initial learning rate of $1e^{-3}$. The learning rate is then annealed following a cosine annealing schedule from $1e^{-3}$ to $1.25e^{-6}$ over the first 50K iterations. We also apply a temperature annealing schedule for the Gumbel-Softmax, defined as $\tau = \text{exp}(-10^{-5} \cdot t)$, where $\tau$ represents the temperature and $t$ is the global training step. Model and dataset specific hyperparameters are detailed in Section~\ref{sec:hyperparameters}.

\subsection{Datasets}\label{sec:datasets}

\begin{table}[t!]
    \caption{Hyperparameters of the best performing models.}
    \label{tab:hyperparameters}
    \centering
    \scriptsize
    \begin{tabular}{lcccc}
        \toprule
        model & Shapes3D & Isaac3D & MPI3D \\
        \midrule
        $\beta$-VAE & $\beta = 10^{-4}$ & $\beta = 5 \times 10^{-5}$ & $\beta = 10^{-5}$ \\
        \cmidrule{2-4}
        \multirow{2}{*}{FactorVAE} & $\beta = 10^{-4}$ & $\beta = 5 \times 10^{-5}$ & $\beta = 10^{-5}$ \\
        & $\gamma = 10^{-4}$ & $\gamma = 5 \times 10^{-5}$ & $\gamma = 10^{-5}$ \\
        \cmidrule{2-4}
        \multirow{2}{*}{BioAE} & $\beta_{\text{nonneg}} = 1$ & $\beta_{\text{nonneg}} = 1$ & $\beta_{\text{nonneg}} = 1$ \\
        & $\beta_{\text{activity}} = 10^{-2}$ & $\beta_{\text{activity}} = 0.1$ & $\beta_{\text{activity}} = 10^{-2}$ \\
        \cmidrule{2-4}    
        $\alpha$-TCVAE & $\alpha = 10^{-5}$ & $\alpha = 10^{-5}$ & $\alpha = 10^{-5}$ \\
        \cmidrule{2-4}        
        \multirow{3}{*}{QLAE} & $\lambda_{\text{quantize}} = 10^{-2}$ & $\lambda_{\text{quantize}} = 10^{-2}$ & $\lambda_{\text{quantize}} = 10^{-2}$ \\
        & $\lambda_{\text{commit}} = 10^{-2}$ & $\lambda_{\text{commit}} = 10^{-2}$ & $\lambda_{\text{commit}} = 10^{-2}$ \\
        & $n_{v} = 16$ & $n_{v} = 16$ & $n_{v} = 16$ \\
        \cmidrule{2-4}
        dVAE & $\beta = 5 \times 10^{-3}$ & $\beta = 10^{-5}$ & $\beta = 5 \times 10^{-5}$ \\
        \cmidrule{2-4}
        QVAE & $\beta = 10^{-3}$ & $\beta = 5 \times 10^{-5}$ & $\beta = 5 \times 10^{-5}$ \\
        \cmidrule{2-4}
        \multirow{2}{*}{FactordVAE} & $\beta = 5 \times 10^{-3}$ & $\beta = 10^{-5}$ & $\beta = 5 \times 10^{-5}$ \\
        & $\gamma = 5 \times 10^{-4}$ & $\gamma = 10^{-5}$ & $\gamma = 5 \times 10^{-6}$ \\
        \cmidrule{2-4}
        \multirow{2}{*}{FactorQVAE} & $\beta = 10^{-3}$ & $\beta = 5 \times 10^{-5}$ & $\beta = 5 \times 10^{-5}$ \\
        & $\gamma = 10^{-4}$ & $\gamma = 5 \times 10^{-5}$ & $\gamma = 10^{-6}$ \\
        \bottomrule
  \end{tabular}
\end{table}

\begin{table}[t]
\centering
\caption{Effect of $\beta$ and $\gamma$ on Isaac3D.}
\label{tab:sensitivity_isaac3d}
\begin{tabular}{cccccc}
\toprule
$\beta$ & $\gamma$ & $\mathsf{D}$ $\uparrow$ & $\mathsf{I}$ $\uparrow$ & $\mathsf{C}$ $\uparrow$ & MSE $\downarrow$ \\
\midrule
$5\!\times\!10^{-5}$ & $5\!\times\!10^{-5}$ & 0.59 & 0.75  & 0.56 & $7\!\times\!10^{-5}$ \\
$1\!\times\!10^{-5}$ & $5\!\times\!10^{-5}$ & 0.56 & 0.74 & 0.50 & $6\!\times\!10^{-5}$  \\
$5\!\times\!10^{-5}$ & $1\!\times\!10^{-5}$ & 0.57 & 0.74 & 0.52 & $6\!\times\!10^{-5}$ \\
$1\!\times\!10^{-5}$ & $1\!\times\!10^{-5}$ & 0.54 & 0.73 & 0.51 & $5\!\times\!10^{-5}$ \\
\bottomrule
\end{tabular}
\end{table}

\textbf{Shapes3D: } It is a dataset consisting 480.000 images of various 3D geometric objects. It has 6 generative factors: floor hue, wall hue, object hue, object scale, object shape, and camera orientation having 10, 10, 10, 8, 4, and 15 possible values, respectively.

\textbf{Isaac3D: } It is a dataset consisting 737.280 images of a synthetic robot arm holding objects in different configurations. It has 9 generative factors: object shape, robot's horizontal axis, robot's vertical axis, camera height, object scale, lighting intensity, lighting direction, object color, and wall color having 3, 8, 5, 4, 4, 4, 6, 4, and 4 possible values, respectively. 

\textbf{MPI3D: } It is a dataset consisting 460.800 images of a real robot arm holding objects in different configurations. It has 7 generative factors: object color, object shape, object size, camera height, background color, robot's horizontal axis, and robot's vertical axis having 4, 4, 2, 3, 3, 40, and 40 possible values, respectively.

\subsection{Hyperparameters}\label{sec:hyperparameters}

\begin{table}[t!]
\centering
    \caption{Notations of network layers used on all models.} 
    \small
    \begin{tabular}{ll}
    \bottomrule
        \textbf{Notation} & \textbf{Description}\\
        \hline
        $\mathrm{Conv}_{n}^{(3\times3)}$
        & 2D Conv layer (out\_ch$=n$, kernel$=3$, stride$=1$, padding$=1$) \\
        \hline
        $\mathrm{Linear}_{n}$ & Linear layer (out\_ch$=n$) \\
        \hline
        $\mathrm{MaxPool}$
        & 2D Max pooling layer (kernel\_size$=2$)\\
        \hline
        $\mathrm{Upsample}$
        & 2D upsampling layer (scale\_factor$=2$)\\
        \hline
        $\mathrm{EncResBlock}_{n}$
        & $3 \times (\mathrm{ReLU}\to\mathrm{Conv}_{n}^{(3\times3)})\to\mathrm{ReLU}\to\mathrm{Conv}_{n}^{(1\times1)}$ $+$ identity \\
        \hline
        $\mathrm{DecResBlock}_{n}$
        & $\mathrm{ReLU}\to\mathrm{Conv}_{n}^{(1\times1)} \to 3 \times (\mathrm{ReLU}\to\mathrm{Conv}_{n}^{(3\times3)})$ $+$ identity \\
        \hline
        $\mathrm{Dense}_{n}$ & $\mathrm{ReLU}\to\mathrm{Linear}_{n}\to\mathrm{ReLU}\to\mathrm{Linear}_{n}\to\mathrm{ReLU}$\\
        \hline
    \end{tabular}
    \label{tab:building_blocks}
\end{table}

We run controlled experiments for all models on all datasets, and report the key hyperparameters of the best performing models in Table~\ref{tab:hyperparameters}. We use the same coefficient names from the original papers, and try to stick to original hyperparameter values from them. As the complexities of the datasets vary a lot, we naturally obtain the best performances with different hypTo examine the effect of $\beta$ and $\gamma$ coefficients in FactorQVAE, we perform a targeted hyperparameter analysis on Isaac3D as in Table~\ref{tab:sensitivity_isaac3d}. As observed in prior disentanglement methods, increasing $\beta$ or $\gamma$ strengthens disentanglement at some cost to reconstruction, while reducing them improves reconstruction quality. For example, with our default setting ($\beta = 5 \times 10^{-5}$, $\gamma = 5 \times 10^{-5}$), FactorQVAE achieves $7 \times 10^{-5}$ MSE, 0.59 disentanglement, and 0.56 completeness. With smaller values ($\beta = 1 \times 10^{-5}$, $\gamma = 1 \times 10^{-5}$), reconstruction improves to $5 \times 10^{-5}$ MSE but disentanglement decreases to 0.54 and completeness to 0.51. Across all tested values, FactorQVAE maintains non-trivial disentanglement ($\geq 0.54$), allowing practitioners to dial performance toward reconstruction or factor separation depending on task priorities.

\subsection{Model Description}\label{sec:model_description}

We use a similar architecture used in \citep{ramesh2021zeroshot}. The common building blocks used in the architecture are given in Table~\ref{tab:building_blocks}. For all datasets, the image size is 64 ($w = h= 64$). We use a codebook $\mathcal{M} \in \mathbb{R}^{64 \times 1}$. Designs of encoding and decoding parts are given as follows:

\textbf{Encoding:} $x\in\mathbb{R}^{w\times h\times3} \to \mathrm{Conv}_{n}^{(3 \times 3)} \to \left[\mathrm{EncResBlock}_{n}\right]_2 \to \mathrm{MaxPool} \to \left[\mathrm{EncResBlock}_{2*n}\right]_2 \to \mathrm{MaxPool} \to \left[\mathrm{EncResBlock}_{4*n}\right]_2 \to \mathrm{Dense}_{m} \to \mathrm{Linear}_{o} \to z_e(x) \in \mathbb{R}^{p \times 1}$

\textbf{Decoding:} $z_q(x) \in \mathbb{R}^{d \times 1} \to \mathrm{Linear}_{\sfrac{w}{4}\times\sfrac{h}{4}\times 16} \to \left[\mathrm{DecResBlock}_{4*n}\right]_2 \to \mathrm{UpSample} \to \left[\mathrm{DecResBlock}_{2*n}\right]_2 \to \mathrm{UpSample} \to \left[\mathrm{DecResBlock}_{n}\right]_2 \to \mathrm{ReLU}\to\mathrm{Conv}_{3}^{(1\times1)} \to \hat{x} \in \mathbb{R}^{w \times h \times 3}$

$n = 128$, and $m = 128$ for all datasets. For AE, BioAE, QLAE, QVAE, and FactorQVAE, $o = p = d$ where $d$ is the latent dimension. For $\beta$-VAE, FactorVAE, and $\alpha$-TCVAE, $x$ is encoded into a latent representation with $o = 2 \times d$ dimensions for the mean and the variance of the posterior distribution, and $p = d$. For dVAE and FactordVAE, $o = p = d*K$ as the encoding part learns the parameters of a distribution over the codebook elements. 

\bibliographystyle{elsarticle-num-names}
\bibliography{bibfile}

\end{document}